\documentclass[11pt]{article}
\usepackage{t1enc}
\usepackage{times}
\usepackage{fullpage}     
\usepackage{amsfonts}

\newcommand{\set}[1]{\left\{ #1\right\}}
\newcommand{\gilt}{:}
\newcommand{\sodass}{\,:\,}
\newcommand{\setGilt}[2]{\left\{ #1\sodass #2\right\}}

\newcommand{\realrange}[2]{\left[#1, #2\right]}

\newcommand{\unitrange}[2]{\realrange{0}{1}}

\newcommand{\llabel}[1]{\label{\labelprefix:#1}}
\newcommand{\labelprefix}{} 

\newcommand{\discussionsize}{\small}

\newcommand{\frage}[1]{}

\newenvironment{code}{\noindent
\begin{tabbing}%
\hspace{2em}\=\hspace{2em}\=\hspace{2em}\=\hspace{2em}\=\hspace{2em}\=%
\hspace{2em}\=\hspace{2em}\=\hspace{2em}\=\hspace{2em}\=\hspace{2em}\=%
\kill}{\end{tabbing}}

\newcommand{\labelcommand}{}
\newcommand{\captiontext}{}
\newsavebox{\codeparam}
\newcounter{lineNumber}
\newenvironment{disscodepos}[3]{%
\renewcommand{\labelcommand}{#2}%
\renewcommand{\captiontext}{#3}%
\sbox{\codeparam}{\parbox{\textwidth}{#3}}%
\begin{figure}[#1]\begin{center}\begin{code}\setcounter{lineNumber}{1}}{%
\end{code}\end{center}\caption{\llabel{\labelcommand}\captiontext}\end{figure}}

{\end{disscodepos}}

\newcommand{\Is}       {:=}

\newdimen\endofsize\endofsize=0.5em
\def\endofbeweis{~\quad\hglue\hsize minus\hsize
                 \hbox{\vrule height \endofsize width
\endofsize}\par}

\usepackage{epsfig}
\usepackage{url}
\usepackage{amsmath}
\usepackage{amssymb}
\usepackage{pdflscape}
\usepackage{latexsym}
\usepackage{todonotes}
\usepackage{algorithmic}
\usepackage{algorithm}
\usepackage{multicol}
\usepackage{wrapfig}
\usepackage{numprint}
\npdecimalsign{.} 
\usepackage{theorem}
\usepackage{makeidx}

\usepackage[left=2cm,top=1.5cm,right=2cm]{geometry}
\def\MdR{\ensuremath{\mathbb{R}}}

\newcommand{\expansion}{\mathrm{expansion}}

\newcommand{\Id}[1]{\ensuremath{\mathit{#1}}}
\usepackage{color}

\newcommand{\mytitle}{Distributed Evolutionary Graph Partitioning}
\begin{document}
\title{\mytitle}
\author{Peter Sanders, Christian Schulz\\ 
	\textit{Karlsruhe Institute of Technology},
	\textit{Karlsruhe, Germany} \\
	\textit{Email: \{\url{sanders}, \url{christian.schulz}\}\url{@kit.edu}} }
\date{}

\maketitle
\begin{abstract}
We present a novel distributed evolutionary algorithm, KaFFPaE, to solve the Graph Partitioning Problem, which makes use of KaFFPa (Karlsruhe Fast Flow Partitioner). The use of our multilevel graph partitioner KaFFPa provides new effective crossover and mutation operators.  By combining these with a scalable communication protocol we obtain a system that is able to improve the best known partitioning results for many inputs in a very short amount of time. 
For example, in Walshaw's well known benchmark tables we are able to improve or recompute 76\% of entries for the tables with 1\%, 3\% and 5\% imbalance. 
\end{abstract}
\thispagestyle{empty}
\section{Introduction}
Problems of \textit{graph partitioning} arise in various areas of computer science, engineering, and related fields. 
For example in high performance computing \cite{schloegel2000gph}, community detection in social networks \cite{journals/corr/abs-0905-4918} and route planning \cite{journals/jea/BauerDSSSW10}. 
In particular the graph partitioning problem is very valuable for parallel computing.
In this area, graph partitioning is mostly used to partition the underlying graph model of computation and communication.
Roughly speaking, vertices in this graph represent computation units and edges denote communication. 
This graph needs to be partitioned such that there are few edges between the blocks (pieces). 
In particular, if we want to use $k$ processors we want to partition the graph into $k$ blocks of about equal size. 

In this paper we focus on a version of the problem that constrains the maximum block size to $(1+\epsilon)$ times the average block size and tries to minimize the total cut size, i.e., the number of edges that run between blocks.
It is well known that this problem is NP-complete \cite{journals/ipl/BuiJ92} and that there is no approximation algorithm with a constant ratio factor for general graphs \cite{journals/ipl/BuiJ92}. Therefore mostly heuristic algorithms are used in practice.  

\begin{wrapfigure}{r}{7cm}
\begin{center}
\vspace*{-0.4cm}
\includegraphics[width=0.4\textwidth]{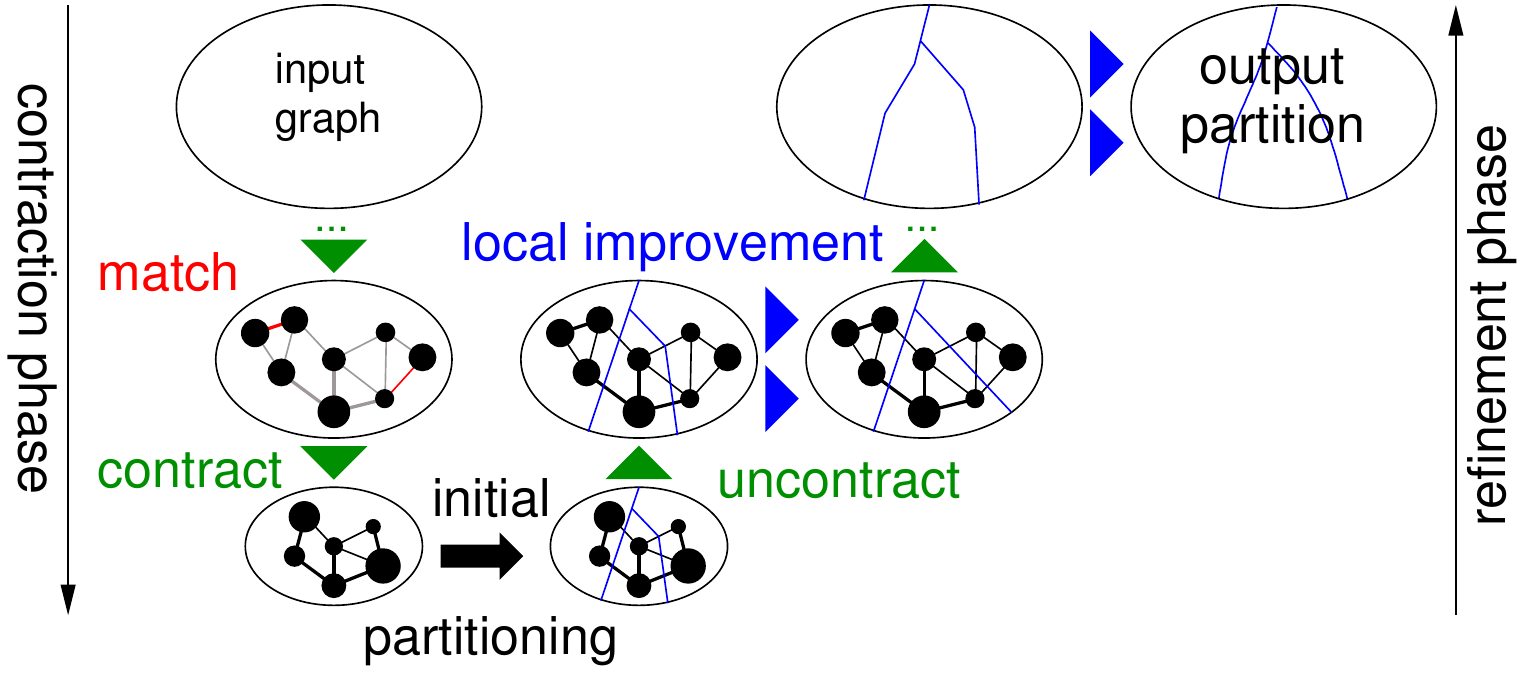}
\end{center}
\label{fig:mgp}
\vspace*{-.75cm}
\caption{Multilevel graph partitioning.}
\end{wrapfigure}

A successful heuristic for partitioning large graphs is the \emph{multilevel graph partitioning} (MGP) approach depicted in Figure~\ref{fig:mgp}
where the graph is recursively \emph{contracted} to achieve smaller graphs which should reflect the same basic structure as the input graph. After applying an \emph{initial partitioning} algorithm to the smallest graph, the contraction is undone and, at each level, a
\emph{local refinement} method is used to improve the partitioning induced by the coarser level. 

The main focus of this paper is a technique which integrates an evolutionary search algorithm with our multilevel graph partitioner KaFFPa and its scalable parallelization. 
We present novel mutation and combine operators which in contrast to previous methods that use a graph partitioner \cite{soper2004combined, delling2010graph} do not need random perturbations of edge weights. 
We show in Section~\ref{s:experiments} that the usage of edge weight perturbations decreases the overall quality of the underlying graph partitioner. 
The new combine operators enable us to combine individuals of different kinds (see Section~\ref{s:evolutionarycomponents} for more details).
Due to the parallelization our system is able to compute partitions that have quality comparable or better than previous entries in Walshaw's well known partitioning benchmark \textit{within a few minutes} for graphs of moderate size. 
Previous methods of Soper et.al \cite{soper2004combined} required runtimes of up to one week for graphs of that size.
We therefore believe that in contrast to previous methods, our method is very valuable in the area of high performance computing.

The paper is organized as follows.  
We begin in Section~\ref{s:preliminaries} by introducing basic concepts. 
After shortly presenting Related Work in Section~\ref{s:related}, we continue describing the main evolutionary components in Section~\ref{s:evolutionarycomponents} and its 
parallelization in Section~\ref{s:parallelization}. 
A summary of extensive experiments done to tune the algorithm and evaluate its performance is presented in Section~\ref{s:experiments}. 
A brief outline of the techniques used in the multilevel graph partitioner KaFFPa is provided in Appendix~\ref{s:kaffpa}. 
We have implemented these techniques in the graph partitioner KaFFPaE (Karlsruhe Fast Flow Partitioner Evolutionary) which is written in C++. 
Experiments reported in Section~\ref{s:experiments} indicate that KaFFPaE is able to compute partitions of very high quality and scales well to large networks and machines. 
\section{Preliminaries}\label{s:preliminaries}
\subsection{Basic concepts}
Consider an undirected graph $G=(V,E,c,\omega)$ 
with edge weights $\omega: E \to \MdR_{>0}$, node weights
$c: V \to \MdR_{\geq 0}$, $n = |V|$, and $m = |E|$.
We extend $c$ and $\omega$ to sets, i.e.,
$c(V')\Is \sum_{v\in V'}c(v)$ and $\omega(E')\Is \sum_{e\in E'}\omega(e)$.
$\Gamma(v)\Is \setGilt{u}{\set{v,u}\in E}$ denotes the neighbors of $v$.
We are looking for \emph{blocks} of nodes $V_1$,\ldots,$V_k$ 
that partition $V$, i.e., $V_1\cup\cdots\cup V_k=V$ and $V_i\cap V_j=\emptyset$
for $i\neq j$. The \emph{balancing constraint} demands that 
$\forall i\in \{1..k\}\gilt c(V_i)\leq L_{\max}\Is (1+\epsilon)c(V)/k+\max_{v\in V} c(v)$ for
some parameter $\epsilon$. 
The last term in this equation arises because each node is atomic and
therefore a deviation of the heaviest node has to be allowed.
The objective is to minimize the total \emph{cut} $\sum_{i<j}w(E_{ij})$ where 
$E_{ij}\Is\setGilt{\set{u,v}\in E}{u\in V_i,v\in V_j}$. 
A clustering is also a partition of the nodes, however $k$ is usually not given in advance and the balance constraint is removed. 
A vertex $v \in V_i$ that has a neighbor $w \in V_j, i\neq j$, is a boundary vertex. 
An abstract view of the partitioned graph is the so called \emph{quotient graph}, where vertices represent blocks and edges are induced by connectivity between blocks. 
Given two clusterings $\mathcal{C}_1$ and $\mathcal{C}_2$ the \emph{overlay clustering} is the clustering where each block corresponds to a connected component of the graph $G_\mathcal{E} = (V,E\backslash \mathcal{E})$ where $\mathcal{E}$ is the union of the cut edges of $\mathcal{C}_1$ and $\mathcal{C}_2$, i.e. all edges that run between blocks in either $\mathcal{C}_1$ or $\mathcal{C}_2$.
By default, our initial inputs will have unit edge and node weights. 
However, even those will be translated into weighted problems in the course of the algorithm.

A matching $M\subseteq E$ is a set of edges that do not share any common nodes, i.e., the graph $(V,M)$ has maximum degree one.  \emph{Contracting} an edge $\set{u,v}$ means to replace the nodes $u$ and $v$ by a new node $x$ connected
to the former neighbors of $u$ and $v$. 
We set $c(x)=c(u)+c(v)$ so the weight of a node at each level is the number of nodes it is representing in the original graph. If replacing edges of the form $\set{u,w}$,$\set{v,w}$ would generate two parallel edges $\set{x,w}$, we insert a single edge with
$\omega(\set{x,w})=\omega(\set{u,w})+\omega(\set{v,w})$.
\emph{Uncontracting} an edge $e$ undos its contraction. 
In order to avoid tedious notation, $G$ will denote the current state of the graph
before and after a (un)contraction unless we explicitly want to refer to 
different states of the graph.
The \textit{multilevel approach} to graph partitioning consists of three main phases.
In the \emph{contraction} (coarsening) phase, we iteratively identify matchings $M\subseteq E$ and contract the edges in $M$. 
Contraction should quickly reduce the size of the input and each computed level should reflect the global structure of the input network. 
Contraction is stopped when the graph is small enough to be directly partitioned using some expensive other algorithm. 
In the \emph{refinement} (or uncoarsening) phase, the matchings are iteratively uncontracted.  
After uncontracting a matching, a refinement algorithm moves nodes between blocks in order to improve the cut size or balance. 

KaFFPa, which we use as a base case partitioner, extended the concept of \emph{iterated multilevel algorithms} which was introduced by \cite{walshaw2004multilevel}. 
The main idea is to iterate the coarsening and uncoarsening phase.
Once the graph is partitioned, edges that are between two blocks are not contracted. 
An \emph{F-cycle} works as follows: on \emph{each} level we perform at most \emph{two recursive calls} using different random seeds during contraction and local search.  
A second recursive call is only made the second time that the algorithm reaches a particular level. 
As soon as the graph is partitioned, edges that are between blocks are not contracted.  
This ensures nondecreasing quality of the partition since our refinement algorithms guarantee no worsening and break ties randomly. These so called \textit{global search strategies} are more effective than plain restarts of the algorithm.
\emph{Extending this idea} will yield the new combine and mutation operators described in Section~\ref{s:evolutionarycomponents}.

\noindent Local search algorithms find good solutions in a very short amount of time but often get stuck in local optima. In contrast to local search algorithms, genetic/evolutionary algorithms are good at searching the problem space globally. 
However, genetic algorithms lack the ability of fine tuning a solution, so that local search algorithms can help to improve the performance of a genetic algorithm. 
The combination of an evolutionary algorithm with a local search algorithm is called \textit{hybrid} or \textit{memetic} evolutionary algorithm \cite{conf/gecco/KimHKM11}.

\section{Related Work}\label{s:related}
There has been a huge amount of research on graph partitioning so that we refer the reader to \cite{fjallstrom1998agp,Walshaw07} for more material on multilevel graph partitioning and to \cite{conf/gecco/KimHKM11} for more material on genetic approaches for graph partitioning.
All general purpose methods that are able to obtain good partitions for large real world graphs are based on the multilevel principle outlined in Section~\ref{s:preliminaries}. 
Well known software packages based on this approach include, Jostle~\cite{Walshaw07}, Metis \cite{karypis1999pmk}, and Scotch \cite{Scotch}.  
KaFFPa \cite{kappa} is a MGP algorithm using local improvement algorithms that are based on flows and more localized FM searches. 
It obtained the best results for many graphs in \cite{soper2004combined}. 
Since we use it as a base case partitioner it is described in more detail in Appendix \ref{s:kaffpa}.
KaSPar \cite{kaspar} is a graph partitioner based on the central idea to (un)contract only a single edge between two levels. 
KaPPa \cite{kappa} is a "classical" matching based MGP algorithm designed for scalable parallel execution. 

Soper et al. \cite{soper2004combined} provided the first algorithm that combined an evolutionary search algorithm with a multilevel graph partitioner. Here crossover and mutation operators have been used to compute edge biases, which yield hints for the underlying multilevel graph partitioner.  
Benlic et al. \cite{conf/ieeeconftoolsartintell/benlichao2010} provided a multilevel memetic algorithm for balanced graph partitioning. This approach is able to compute many entries in Walshaw's Benchmark Archive \cite{soper2004combined} for the case $\epsilon=0$. PROBE \cite{journals/tc/ChardaireBM07} is a meta-heuristic which can be viewed as a genetic algorithm without selection. It outperforms other metaheuristics, but it is restricted to the case $k=2$ and $\epsilon=0$.

Very recently an algorithm called PUNCH \cite{delling2010graph} has been introduced. 
This approach is not based on the multilevel principle.  
However, it creates a coarse version of the graph based on the notion of natural cuts. 
Natural cuts are relatively sparse cuts close to denser areas. 
They are discovered by finding minimum cuts between carefully chosen regions of the graph. 
They introduced an evolutionary algorithm which is similar to Soper et al. \cite{soper2004combined}, i.e. using a combine operator that computes edge biases yielding hints for the underlying graph partitioner. 
Experiments indicate that the algorithm computes very good partitions for road networks.  
For instances without a natural structure such as road networks, natural cuts are not very helpful. 

\section{Evolutionary Components} \label{s:evolutionarycomponents}
The general idea behind evolutionary algorithms (EA) is to use mechanisms which are highly inspired by biological evolution such as selection, mutation, recombination and survival of the fittest. 
An EA starts with a population of individuals (in our case partitions of the graph) and evolves the population into different populations over several rounds. 
In each round, the EA uses a selection rule based on the fitness of the individuals (in our case the edge cut) of the population to select good individuals and combine them to obtain improved offspring \cite{goldbergGA89}. 
Note that we can use the cut as a fitness function since our partitioner almost always generates partitions that are within the given balance constraint, i.e. there is no need to use a penalty function or something similar to ensure that the final partitions generated by our algorithm are feasible. 
When an offspring is generated an eviction rule is used to select a member of the population and replace it with the new offspring. 
In general one has to take both into consideration, the fitness of an individual and the distance between individuals in the population \cite{baeckEvoAlgPHD96}. 
Our algorithm generates only one offspring per generation. Such an evolutionary algorithm is called \textit{steady-state} \cite{dejongEvoComp2006}. 
A typical structure of an evolutionary algorithm is depicted in Algorithm~\ref{alg:generalsteadystateEA}.

For an evolutionary algorithm it is of major importance to keep the diversity in the population high \cite{baeckEvoAlgPHD96}, i.e. the individuals should not become too similar, in order to avoid a premature convergence of the algorithm.  
In other words, to avoid getting stuck in local optima a procedure is needed that randomly perturbs the individuals.
In classical evolutionary algorithms, this is done using a mutation operator. 
It is also important to have operators that introduce unexplored search space to the population. 
Through a new kind of crossover and mutation operators, introduced in Section~\ref{s:combineoperators}, we introduce more elaborate diversification strategies which allow us to search the search space more effectively. 

Interestingly, Inayoshi et al. \cite{conf/ppsn/InayoshiM94} noticed that good local solutions of the graph partitioning problem tend to be close to one another. 
Boese et al. \cite{boese1994new} showed that the quality of the local optima overall decreases as the distance from the global optimum increases. 
We will see in the following that our combine operators can exchange good parts of solutions quite effectively especially if they have a small distance.  

\begin{center}
\small
\begin{algorithm}[h!]
\begin{algorithmic}
\STATE \textbf{procedure} \textit{steady-state-EA}
\STATE   \quad create initial population $P$ 
\STATE   \quad \textbf{while} stopping criterion not fulfilled 
\STATE   \quad \quad \textit{select} parents $p_1, p_2$ from $P$ 
\STATE   \quad \quad \textit{combine} $p_1$ with $p_2$ to create offspring $o$
\STATE   \quad \quad \textit{mutate} offspring $o$ 
\STATE   \quad \quad \textit{evict} individual in population using $o$ 
\STATE   \quad \textbf{return} the fittest individual that occurred
\end{algorithmic}
\caption{A classic general steady-state evolutionary algorithm.}
\label{alg:generalsteadystateEA}
\end{algorithm}
\end{center}

\subsection{Combine Operators} \label{s:combineoperators}
We now describe the general combine operator framework. This is followed by three instantiations of this framework. 
In contrast to previous methods that use a multilevel framework our combine operators do not need perturbations of edge weights since we integrate the operators into our partitioner and do not use it as a complete black box.

Furthermore all of our combine operators assure that the offspring has a partition quality \textit{at least as good as the best of both parents}.  
Roughly speaking, the combine operator framework combines an individual/partition $\mathcal{P} = V^\mathcal{P}_1, ..., V^\mathcal{P}_k$ (which has to fulfill a balance constraint) with a clustering $\mathcal{C} = V^\mathcal{C}_1, ..., V^\mathcal{C}_{k'}$. 
Note that
\begin{wrapfigure}{r}{6.1cm}
\begin{center}
\vspace*{-0.5cm}
\includegraphics[width=3.5cm]{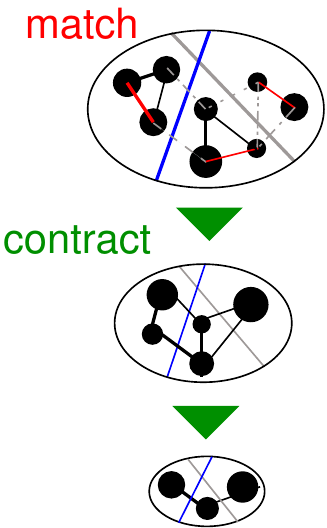}
\end{center}
\caption{On the top a graph $G$ with two partitions, the dark and the light line, are shown. Cut edges are not eligible for the matching algorithm. Contraction is done until no matchable edge is left. The best of the two given partitions is used as initial partition.}
\vspace*{-0.5cm}
\label{fig:generalcrossover}
\end{wrapfigure}
the clustering does not necessarily has to fulfill a balance constraint and $k'$ is not necessarily given in advance. 
All instantiations of this framework use a different kind of clustering or partition.
The partition and the clustering are both used as input for our multi-level graph partitioner KaFFPa in the following sense. 
Let $\mathcal{E}$ be the set of edges that are cut edges, i.e. edges that run between two blocks, in either $\mathcal{P}$ \textit{or} $\mathcal{C}$. 
All edges in $\mathcal{E}$ are blocked during the coarsening phase, i.e. they \textit{are not contracted} during the coarsening phase.
In other words these edges are not eligible for the matching algorithm used during the coarsening phase and therefore are not part of any matching computed.
An illustration of this can be found in Figure~\ref{fig:generalcrossover}. 

\begin{figure}[t]
\begin{center}
\begin{tabular}{cc}
\includegraphics[width=4cm]{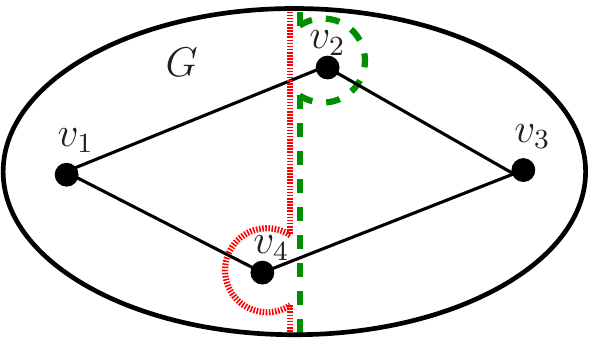}& \quad  \quad \includegraphics[width=4cm]{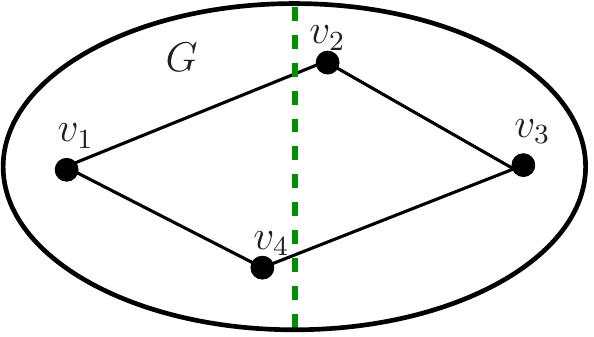}
\end{tabular}
\end{center}
\caption{A graph $G$ and two bipartitions; the dotted and the dashed line (left). Curved lines represent a large cut. The four vertices correspond to the coarsest graph in the multilevel procedure. Local search algorithms can effectively exchange $v_2$ or $v_4$ to obtain the better partition depicted on the right hand side (dashed line).  }
\label{fig:crossover}
\end{figure}

The stopping criterion for the multi-level partitioner is modified such that it stops when no contractable edge is left. 
Note that the coarsest graph is now exactly the same as the quotient graph $\mathcal{Q'}$ of the  overlay clustering of $\mathcal{P}$ and $\mathcal{C}$ of $G$ (see Figure~\ref{fig:crossover}). 
Hence vertices of the coarsest graph correspond to the connected components of $G_\mathcal{E} = (V, E\backslash \mathcal{E})$ and the weight of the edges between vertices corresponds to the sum of the edge weights running between those connected components in $G$.

As soon as the coarsening phase is stopped, we apply the partition $\mathcal{P}$  to the coarsest graph and use this as 
initial partitioning. 
This is possible since we did not contract any cut edge of $\mathcal{P}$. 
Note that due to the specialized coarsening phase and this specialized initial partitioning we obtain a high quality initial solution on a very coarse graph which is usually not discovered by conventional partitioning algorithms. 
Since our refinement algorithms guarantee no worsening of the input partition and use random tie breaking we can assure nondecreasing partition quality.
Note that the refinement algorithms can effectively exchange good parts of the solution on the coarse levels by moving only a few vertices. 
Figure~\ref{fig:crossover} gives an example. 

Also note that this combine operator can be extended to be a multi-point combine operator, i.e. the operator would use $p$ instead of two parents. 
However, during the course of the algorithm a sequence of two point combine steps is executed which somehow "emulates" a multi-point combine step. 
Therefore, we restrict ourselves to the case $p=2$.
When the offspring is generated we have to decide which solution should be evicted from the current population.
We evict the solution that is \textit{most similar} to the offspring among those individuals in the population that have a cut worse or equal than the offspring itself. 
The difference of two individuals is defined as the size of the symmetric difference between their sets of cut edges. 
This ensures some diversity in the population and hence makes the evolutionary algorithm more effective.

\subsubsection{Classical Combine using Tournament Selection}
This instantiation of the combine framework corresponds to a classical evolutionary combine operator $C_1$. 
That means it takes two individuals $P_1, P_2$ of the population and performs the combine step described above.
In this case $\mathcal{P}$ corresponds to the partition having the smaller cut and $\mathcal{C}$ corresponds to the partition having the larger cut. 
Random tie breaking is used if both parents have the same cut.
The selection process is based on the tournament selection rule \cite{Miller95geneticalgorithms}, i.e. $P_1$ is the fittest out of two random individuals $R_1, R_2$ from the population. 
The same is done to select $P_2$. 
Note that in contrast to previous methods the generated offspring will have a cut smaller or equal to the cut of $\mathcal{P}$.
Due to the fact that our multi-level algorithms are randomized, a combine operation performed twice using the same parents can yield different offspring. 
\subsubsection{Cross Combine / (Transduction)}
In this instantiation of the combine framework $C_2$, the clustering $\mathcal{C}$ corresponds to a partition of $G$. 
But instead of choosing an individual from the population we create a new individual in the following way. 
We choose $k'$ uniformly at random in $[k/4, 4k]$ and $\epsilon'$ uniformly at random in $[\epsilon, 4\epsilon]$. 
We then use KaFFPa to create a $k'$-partition of $G$ fulfilling the balance constraint $\max c(V_i) \leq (1+\epsilon')c(V)/k'$.
In general larger imbalances reduce the cut of a partition which then yields good clusterings for our crossover. 
To the best of our knowledge there has been no genetic algorithm that performs combine operations combining individuals from different search spaces.

\subsubsection{Natural Cuts} \label{s:combinenaturalcut}
Delling et al. \cite{delling2010graph} introduced the notion of \textit{natural cuts} as a preprocessing technique for the partitioning of road networks. 
The preprocessing technique is able to find relatively  sparse cuts close to denser areas.
We use the computation of natural cuts to provide another combine operator, i.e. combining a $k$-partition with a clustering generated by the computation of natural cuts. 
We closely follow their description: 
The computation of natural cuts works in rounds. 
Each round picks a center vertex $v$ and grows a breadth-first search (BFS) tree. 
The BFS is stopped as soon as the weight of the tree, i.e. the sum of the vertex weights of the tree, reaches \begin{wrapfigure}{r}{.3\textwidth}
\begin{center}
\includegraphics[width=3cm]{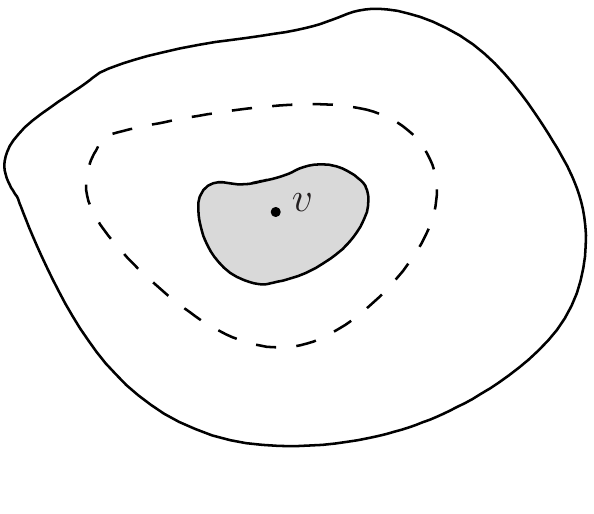} \\  \includegraphics[width=5cm]{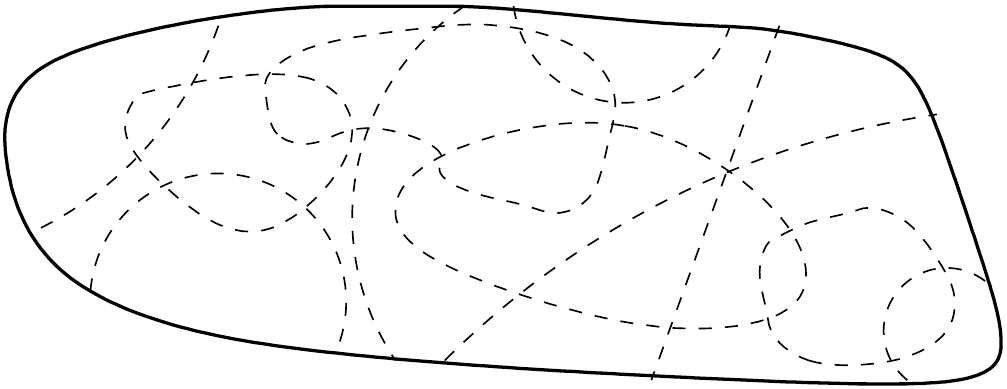}
\end{center}
\caption{On the top we see the computation of a natural cut. A BFS Tree which starts from $v$ is grown. The gray area is the core. The dashed line is the natural cut. It is the minimum cut between the contracted versions of the core and the ring (shown as the solid line).  During the computation several natural cuts are detected in the input graph (bottom).}

\vspace*{-0.5cm}
\label{fig:naturalcutsexplained}
\end{wrapfigure}
$\alpha U$, for some parameters $\alpha$ and $U$. The set of the neighbors of $T$ in  $V \backslash T$ is called the \textit{ring} of $v$. 
The \textit{core} of $v$ is the union of all vertices added to $T$ before its size reached $\alpha U / f$ where $f > 1$ is another parameter.

The core is then temporarily contracted to a single vertex $s$ and the ring into a single vertex $t$ to compute the minimum $s$-$t$-cut between them using the given edge weights as capacities. 

To assure that every vertex eventually belongs to at least one core, and therefore is inside at least one cut, the vertices $v$ are picked uniformly at random among all vertices that have not yet been part of any core in any round.
The process is stopped when there are no such vertices left. 

In the original work \cite{delling2010graph} each connected component of the graph $G_C = (V, E \backslash C)$, where $C$ is the union of all edges cut by the process above, is contracted to a single vertex. 
Since we do not use natural cuts as a preprocessing technique at this place we don't contract these components. 
Instead we build a clustering $\mathcal{C}$ of $G$ such that each connected component of $G_C$ is a block.

This technique yields the third instantiation of the combine framework $C_3$ which is divided into two stages, i.e. the clustering used for this combine step is dependent on the stage we are currently in. 
In both stages the partition $\mathcal{P}$ used for the combine step is selected from the population using tournament selection. 
During the first stage we choose $f$ uniformly at random in $[5,20]$, $\alpha$ uniformly at random in $[0.75, 1.25]$ and we set $U =  |V|/3k$.
Using these parameters we obtain a clustering $\mathcal{C}$ of the graph which is then used in the combine framework described above. 
This kind of clustering is used until we reach an upper bound of ten calls to this combine step.
When the upper bound is reached we switch to the second stage.
In this stage we use the clusterings computed during the first stage, i.e. we extract elementary natural cuts and use them to quickly compute new clusterings.
An \textit{elementary natural cut} (ENC) consists of a set of cut edges and the set of nodes in its core. 
Moreover, for each node $v$ in the graph, we store the set of of ENCs $N(v)$ that contain $v$ in their core. 
With these data structures its easy to pick a new clustering $\mathcal{C}$ (see Algorithm \ref{alg:computeNCclustering}) which is then used in the combine framework described above.

\begin{algorithm}[h!]
\small
        \begin{algorithmic}[1]
        \STATE unmarked all nodes in $V$ 
        \STATE \textbf{for each} $v \in V$ in random order \textbf{do}
        \STATE \quad \textbf{if} $v$ is not marked \textbf{then}
        \STATE \quad \quad    pick a random ENC $C$ in $N(v)$
        \STATE \quad \quad    output $C$
        \STATE \quad \quad    mark all nodes in $C$'s core
\end{algorithmic}
\caption{computeNaturalCutClustering (second stage)}
\label{alg:computeNCclustering}
\end{algorithm}

\subsection{Mutation Operators}
We define two mutation operators, an ordinary and a modified F-cycle. 
Both mutation operators use a random individual from the current population.
The main idea is to iterate coarsening and refinement several times using different seeds for random tie breaking. 
The first mutation operator $M_1$ can assure that the quality of the input partition does not decrease. 
It is basically an ordinary F-cycle which is an algorithm used in KaFFPa. 
Edges between blocks are not contracted.
The given partition is then used as initial partition of the coarsest graph.  
In contrast to KaFFPa, we now can use the partition as input to the partition in the very beginning. 
This ensures nondecreasing quality since our refinement algorithms guarantee no worsening. 
The second mutation operator $M_2$  works quite similar with the small difference that the input partition is not used as initial partition of the coarsest graph. 
That means we obtain very good coarse graphs but we can not assure that the final individual has a higher quality than the input individual.
In both cases the resulting offspring is inserted into the population using the eviction strategy described in Section~\ref{s:combineoperators}.
\section{Putting Things Together and Parallelization}
\label{s:parallelization}
We now explain the parallelization and describe how everything is put together. Each processing element (PE)  basically performs the same operations using different random seeds (see Algorithm~\ref{alg:localview}). 
First we estimate the population size $S$: each PE performs a partitioning step and measures the time $\overline{t}$ spend for partitioning. 
We then choose $S$ such that the time for creating $S$ partitions is approximately $t_{\text{total}}/f$ where the fraction $f$ is a tuning parameter and $t_{\text{total}}$ is the total running time that the algorithm is given to produce a partition of the graph. 
Each PE then builds its own population, i.e. KaFFPa is called several times to create $S$ individuals/partitions. 
Afterwards the algorithm proceeds in rounds as long as time is left. 
With corresponding probabilities, mutation or combine operations are performed and the new offspring is inserted into the population.

We choose a parallelization/communication protocol that is quite similar to \textit{randomized rumor spreading} \cite{conf/icalp/DoerrF11}. 
Let $p$ denote the number of PEs used. A communication step is organized in rounds. 
In each round, a PE chooses a communication partner and sends her the currently best partition $P$ of the local population. 
The selection of the communication partner is done uniformly at random among those PEs to which $P$ not already has been send to.  
Afterwards, a PE checks if there are incoming individuals and if so inserts them into the local population using the eviction strategy described above.
If $P$ is improved, all PEs are again eligible.
This is repeated $\log p$ times.
Note that the algorithm is implemented \textit{completely asynchronously}, i.e. there is no need for a global synchronisation.
The process of creating individuals is parallelized as follows:
Each PE makes $s' = |S|/p$ calls to KaFFPa using different seeds to create $s'$ individuals. 
Afterwards we do the following $S-s'$ times:
The root PE computes a random cyclic permutation of all PEs and broadcasts it to all PEs. 
Each PE then sends a random individual to its successor in the cyclic permutation and receives a individual from 
its predecessor in the cyclic permutation. 
We call this particular part of the algorithm \textit{quick start}.

The ratio $\frac{c}{10}:\frac{10-c}{10}$ of mutation to crossover operations yields a tuning parameter $c$. 
As we will see in Section~\ref{s:experiments} the ratio $1:9$ is a good choice.
After some experiments we fixed the ratio of the mutation operators $M_1:M_2$ to $4:1$ and the ratio of the combine operators $C_1:C_2:C_3$ to $3:1:1$.

Note that the communication step in the last line of the algorithm could also be performed only every $x$-iterations (where $x$ is a tuning parameter) to save communication time.
Since the communication network of our test system is very fast (see Section~\ref{s:experiments}), we perform the communication step in each iteration.

\begin{center}
\vspace*{-0.3cm}
\begin{algorithm}[h!]
\small
\begin{algorithmic}
\STATE \textbf{procedure} \textit{locallyEvolve}
\STATE   \quad estimate population size $S$
\STATE   \quad \textbf{while} time left 
\STATE   \quad \quad \textbf{if} elapsed time $< t_{\text{total}}/f$ \textbf{then} create individual and insert into local population 
\STATE   \quad \quad \textbf{else}
\STATE   \quad \quad\quad flip coin $c$ with corresponding probabilities
\STATE   \quad \quad\quad \textbf{if} $c$ shows head \textbf{then}
\STATE   \quad \quad\quad \quad perform a mutation operation  
\STATE   \quad \quad\quad \textbf{else}
\STATE   \quad \quad\quad \quad perform a combine operation 
\STATE   \quad \quad\quad insert offspring into population if possible 
\STATE   \quad \quad communicate according to communication protocol
\end{algorithmic}
\caption{All PEs perform basically the same operations using different random seeds.}
\label{alg:localview}
\end{algorithm}
\vspace*{-0.3cm}
\end{center}

\section{Experiments}\label{s:experiments}
\paragraph*{Implementation.}
We have implemented the algorithm described above using C++.  Overall,
our program (including KaFFPa) consists of about 22\,500 lines of code. 
We use two base case partitioners, KaFFPaStrong and KaFFPaEco. 
KaFFPaEco is a good tradeoff between quality and speed, and KaFFPaStrong is 
focused on quality.
For the following comparisons we used Scotch 5.1.9., and kMetis 5.0 (pre2). 
\paragraph*{System.}
Experiments have been done on two machines. Machine A is a cluster with 200 nodes where each node is equipped with two Quad-core Intel Xeon processors (X5355) which run at a clock speed of 2.667 GHz. 
Each node has 2x4 MB of level 2 cache each and run Suse Linux Enterprise 10 SP 1.  
All nodes are attached to an InfiniBand 4X DDR interconnect which is characterized by its very low latency of below 2 microseconds and a point to point bandwidth between two nodes of more than 1300 MB/s.
Machine B has two Intel Xeon X5550, 48GB RAM, running Ubuntu 10.04. Each CPU has 4 cores (8 cores when hyperthreading is active) running at 2.67 GHz.  
Experiments in Sections \ref{sec:parametertuning}, \ref{sec:expscalability}, \ref{sec:comparisionkaffpaandother} and \ref{sec:walshawbenchmark} have been conducted on machine A, and experiments in Sections \ref{sec:combineopexperiment} and \ref{sec:exproadnetworks} have been conducted on machine B.
All programs were compiled using GCC Version 4.4.3 and optimization level~3 using OpenMPI 1.5.3.
Henceforth, a PE is one core.
\paragraph*{Instances.}
We report experiments on three suites of instances (small, medium sized and road networks) summarized in
Appendix~\ref{sec:instances}. 
\Id{rggX} is a \emph{random geometric graph} with
$2^{X}$ nodes where nodes represent random points in the unit square and edges
connect nodes whose Euclidean distance is below $0.55 \sqrt{ \ln n / n }$.
This threshold was chosen in order to ensure that the graph is almost connected. 
\Id{DelaunayX} is the Delaunay triangulation of $2^{X}$
random points in the unit square.  Graphs \Id{uk},\Id{3elt}..\Id{fe\_body} and
\Id{t60k}..\Id{memplus} come from Walshaw's benchmark archive
\cite{walshaw2000mpm}.  Graphs \Id{deu} and \Id{eur}, \Id{bel} and \Id{nld} are undirected versions of the road networks, used in \cite{DSSW09}. 
\Id{luxemburg} is a road network taken from  \cite{dimacschallengegraphpartandcluster}.
Our default number of partitions $k$ are 2, 4, 8, 16, 32, 64 since they are the default values in \cite{walshaw2000mpm} and in some cases we additionally use 128 and 256. 
Our default value for the allowed imbalance is 3\% since this is one
of the values used in \cite{walshaw2000mpm} and the default value in Metis. 
Our default number of PEs is 16. 

\paragraph*{Methodology.} We mostly present two kinds of data: average values and plots that show the evolution of solution quality (\textit{convergence plots}).  
In both cases we perform multiple repetitions. The number of repetitions is dependent on the test that we perform.
Average values over multiple instances are obtained as follows: for each instance (graph, $k$), we compute the geometric mean of the average edge cut values for each instance. 
We now explain how we compute the convergence plots.
We start explaining how we compute them for a single instance $I$:
whenever a PE creates a partition it reports a pair ($t$, cut), where the timestamp $t$ is the currently elapsed time on the particular PE and cut refers to the cut of the partition that has been created.
When performing multiple repetitions we report average values ($\overline{t}$, avgcut) instead.
After the completion of KaFFPaE we are left with $P$ sequences of pairs ($t$, cut) which we now merge into one sequence.
The merged sequence is sorted by the timestamp $t$. 
The resulting sequence is called $T^I$.
Since we are interested in the evolution of the solution quality, we compute another sequence $T^I_{\text{min}}$.
For each entry (in sorted order) in $T^I$ we insert the entry $(t, \min_{t'\leq t} \text{cut}(t'))$ into $T^I_\text{min}$.
Here $\min_{t'\leq t} \text{cut}(t')$ is the minimum cut that occurred until time $t$.
$N^I_{\text{min}}$ refers to the normalized sequence, i.e. each entry ($t$, cut) in $T^I_\text{min}$ is replaced by ($t_n$, cut) where $t_n = t/t_I$ and $t_I$ is the average time that KaFFPa needs to compute a partition for the instance $I$.
To obtain average values over \textit{multiple instances} we do the following: for each instance we label all entries in $N^I_{\text{min}}$, i.e. ($t_n$, cut) is replaced by ($t_n$, cut, $I$). We then merge all sequences $N^I_\text{min}$ and sort by $t_n$. The resulting sequence is called $S$. 
The final sequence $S_g$ presents \textit{event based} geometric averages values. 
We start by computing the geometric mean cut value $\mathcal{G}$ using the first value of all $N^I_\text{min}$ (over $I$).
To obtain $S_g$ we basically sweep through $S$: for each entry (in sorted order) $(t_n, c, I)$ in $S$ we update $\mathcal{G}$, i.e. the cut value of $I$ that took part in the computation of $\mathcal{G}$ is replaced by the new value $c$, and insert $(t_n, \mathcal{G})$ into $S_g$. 
Note that $c$ can be only smaller or equal to the old cut value of $I$.

\subsection{Parameter Tuning}
\label{sec:parametertuning}
We now tune the fraction parameter $f$ and the ratio between mutation and crossover operations. 
For the parameter tuning we choose our small testset because runtimes for a single graph partitioner call are not too large. 
To save runtime we focus on $k=64$ for tuning the parameters. 
For each instance we gave KaFFPaE ten minutes time and 16 PEs to compute a partition. 
During this test the quick start option is disabled.

\begin{figure}[t!]
\vspace*{-1cm}
\begin{center}
\includegraphics[width=0.4\textwidth]{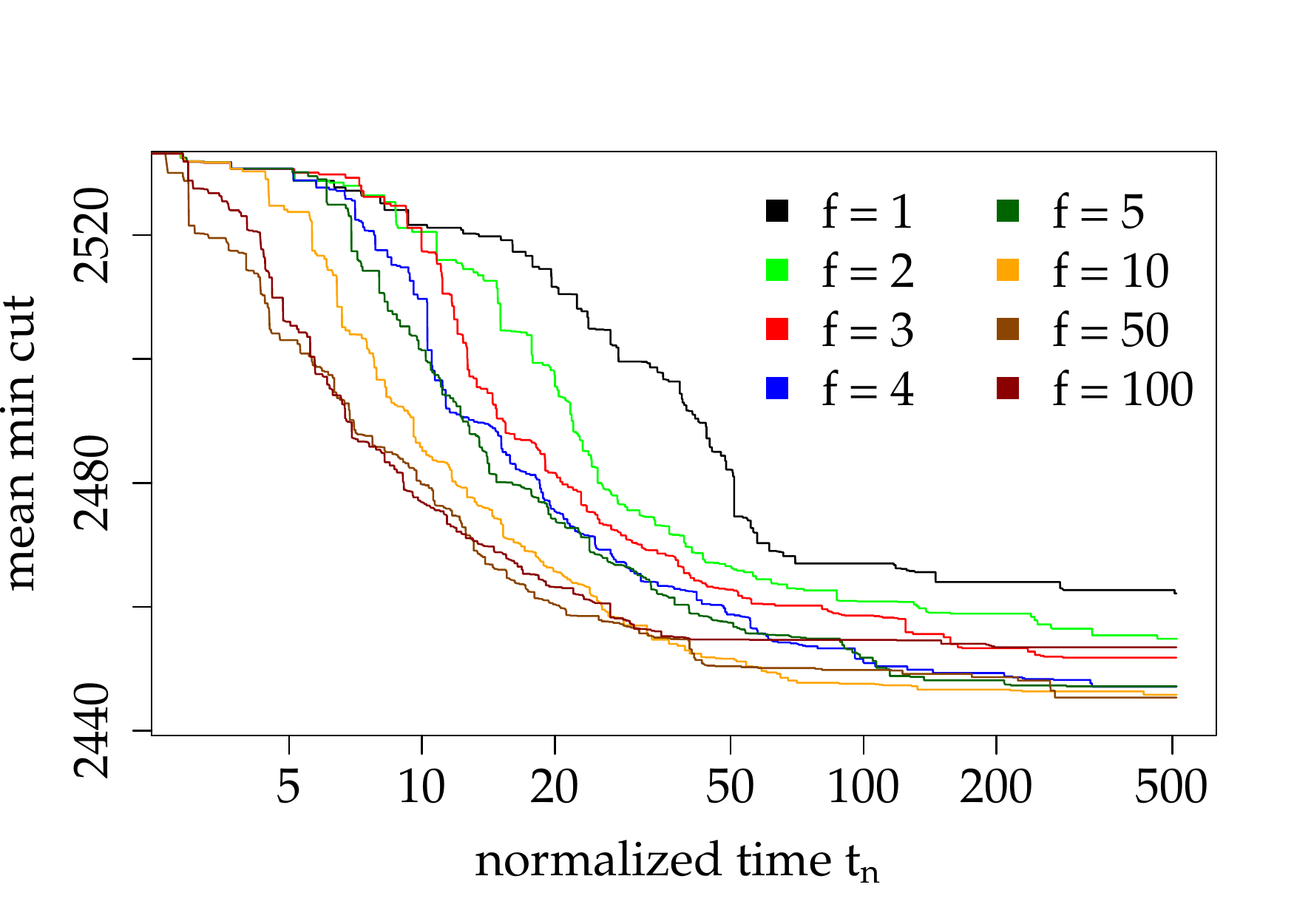}
\includegraphics[width=0.4\textwidth]{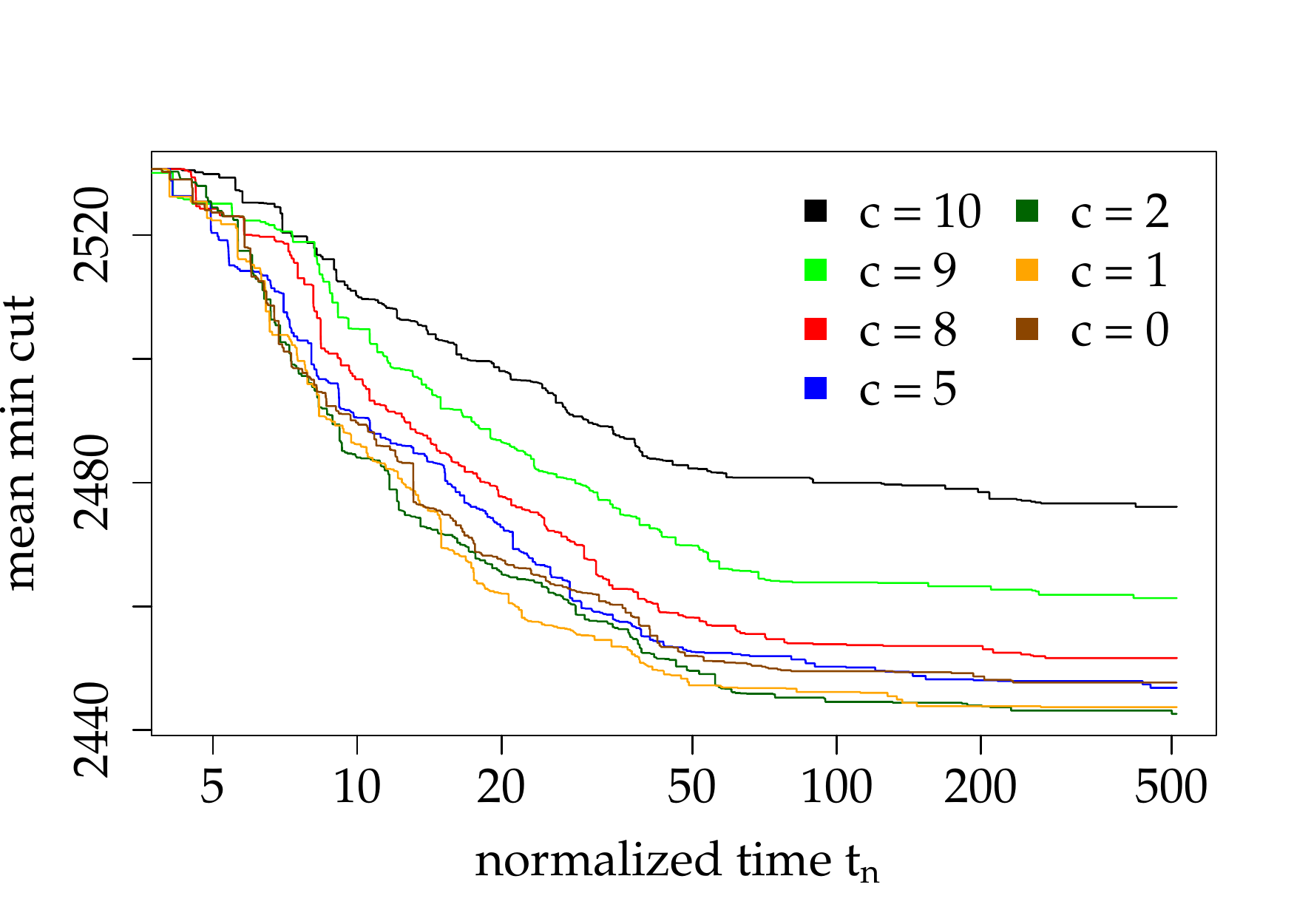}
\end{center}
\vspace*{-1cm}
\caption{Conv. plots for the \textit{fraction} $f$ using $c=1$ (left) and the \textit{flip coin} $c$ using $f=10$ (right). }
\vspace*{-.5cm}

\label{fig:parametertuning}
\end{figure}

For this test the flip coin parameter $c$ is set to one. 
In Figure~\ref{fig:parametertuning} we can see that the algorithm is not too sensitive about the exact choice of this parameter.
However, larger values of $f$ speed up the convergence rate and improve the result achieved in the end. 
Since $f=10$ and $f=50$ are the best parameter in the end, we choose $f=10$ as our default value.
For tuning the ratio $\frac{c}{10}:\frac{10 - c}{10}$ of mutation and crossover operations, we set $f$ to ten.
We can see that for smaller values of $c$ the algorithm is not too sensitive about the exact choice of the parameter. 
However, if the $c$ exceeds 8 the convergence speed slows down which yields worse average results in the end.
We choose $c=1$ because it has a slight advantage in the end.
The parameter tuning uses KaFFPaStrong as a partitioner. 
We also performed the parameter tuning using KaFFPaEco as a partitioner (see Appendix~\ref{sec:furtherparametertuning}).

\subsection{Scalability}
\label{sec:expscalability}
\begin{figure}[h!]
\vspace*{-.5cm}
\begin{center}
\includegraphics[width=5cm]{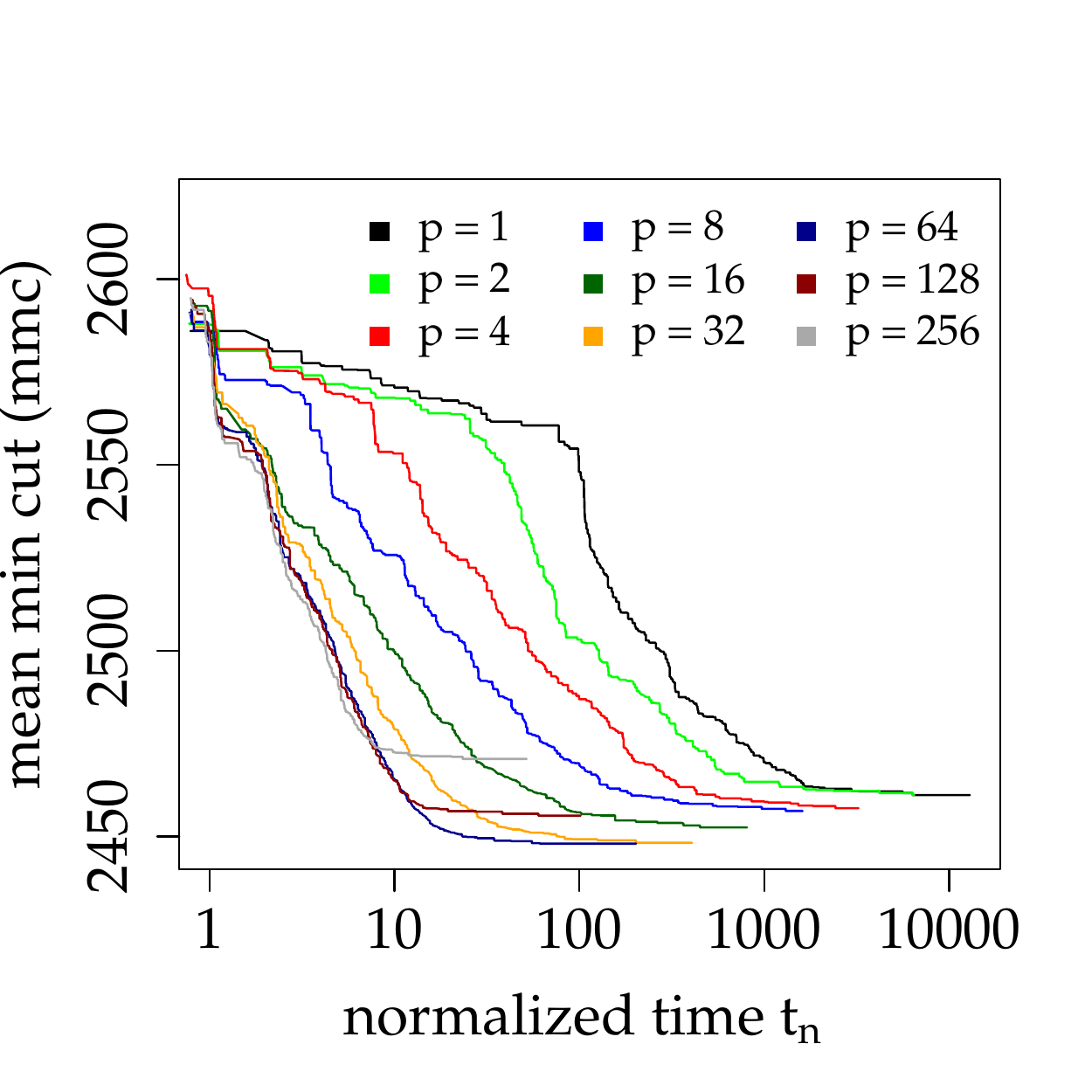} \quad 
\includegraphics[width=5cm]{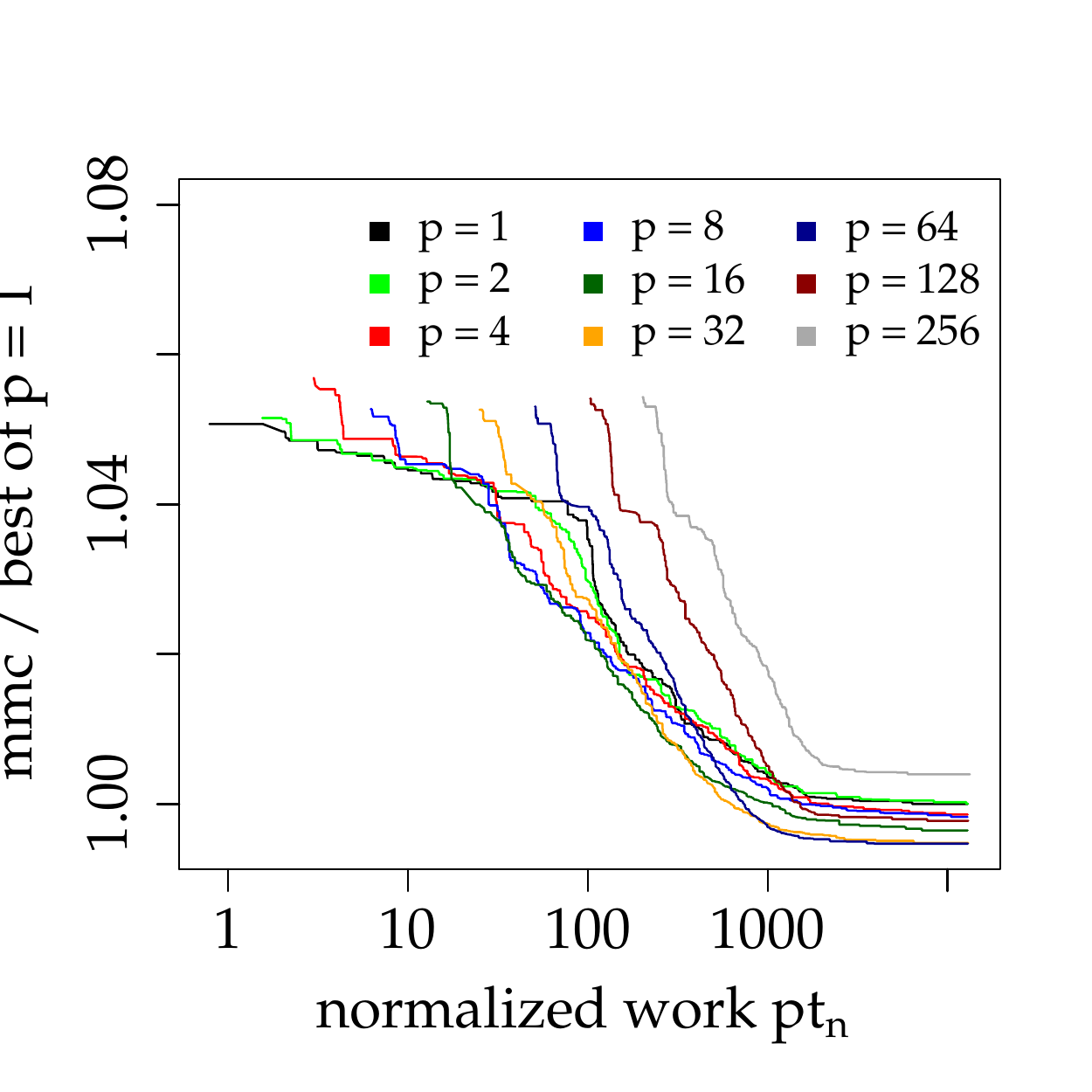} \quad
\includegraphics[width=5cm]{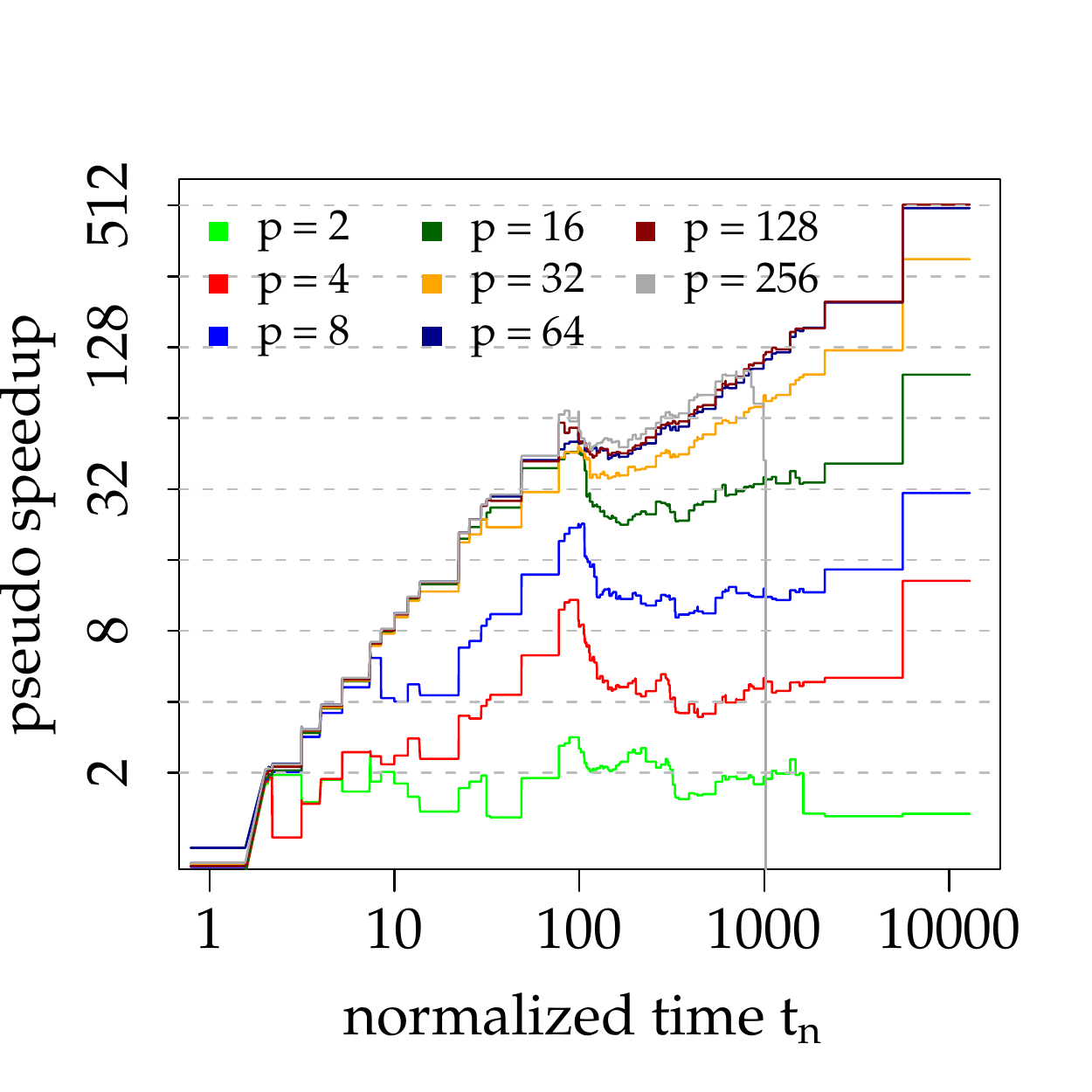}
\end{center}
\vspace*{-0.5cm}
\caption{Scalability of our algorithm: (left) a normal convergence plot, (middle) mean minimum cut relative to best cut of KaFFPaE using one PE, (right) pseudo speedup $S_p(t_n)$ (larger versions can be found in Appendix~\ref{sec:largerscalabilityKaFFPaE}).}
\vspace*{-.5cm}       
\label{fig:scalabilityKaFFPaE}
\end{figure}

In this Section we study the scalability of our algorithm. We do the following to obtain a fair comparison:  
basically each configuration has the same amount of time, i.e. when doubling the number of PEs used, 
we divide the time that KaFFPaE has to compute a partition per instance by two.
To be more precise, when we use one PE KaFFPaE has $t_1=15360s$ to compute a partition of an instance.
When KaFFPaE uses $p$ PEs, then it gets time $t_p=t_1/p$ to compute a partition of an instance. 
For all the following tests the quick start option is enabled.
To save runtime we use our small sized testset and fix $k$ to 64.
Here we perform five repetitions per instance. 
We can see in Figure~\ref{fig:scalabilityKaFFPaE} that using more processors speeds up the convergence speed and up to $p=128$ also \textit{improves} the quality in the end (in these cases the speedups are optimal in the end). 
This might be due to island effects \cite{AlbaT02}.
For $p=256$ results are worse compared to $p=1$. 
This is because the algorithm is barely able to perform combine and mutation steps, due to the very small amount of time given to KaFFPaE (60 seconds). 
On the largest graph of the testset (delaunay16) we need about 20 seconds to create a partition into $k=64$ blocks. 

We now define pseudo speedup $S_p(t_n)$ which is a measure for speedup at a particular normalized time $t_n$ of the configuration using one PE.
Let $c_p(t_n)$ be the mean minimum cut that KaFFPaE has computed using $p$ PEs until normalized time $t_n$.
The pseudo speedup is then defined as $S_p(t_n) = c'_1(t_n)/ c'_p(t_n)$ where $c'_i(t_n) = \min_{c_i(t') \leq c_1(t_n)} t'$. If $c'_p(t) > c'_1(t_n)$ for all $t$ we set $S_p(t_n) = 0$ (in this case the parallel algorithm is not able to compute the result computed by the sequential algorithm at normalized time $t_n$; this is only the case for $p=256$).
We can see in Figure~\ref{fig:scalabilityKaFFPaE} that after a short amount of time we reach super linear pseudo speedups in most cases.

\vspace*{-1cm}
\clearpage
\subsection{Comparison with KaFFPa and other Systems}
\label{sec:comparisionkaffpaandother}
\begin{figure}
\vspace*{-1cm}
\begin{center}
\includegraphics[width=4cm]{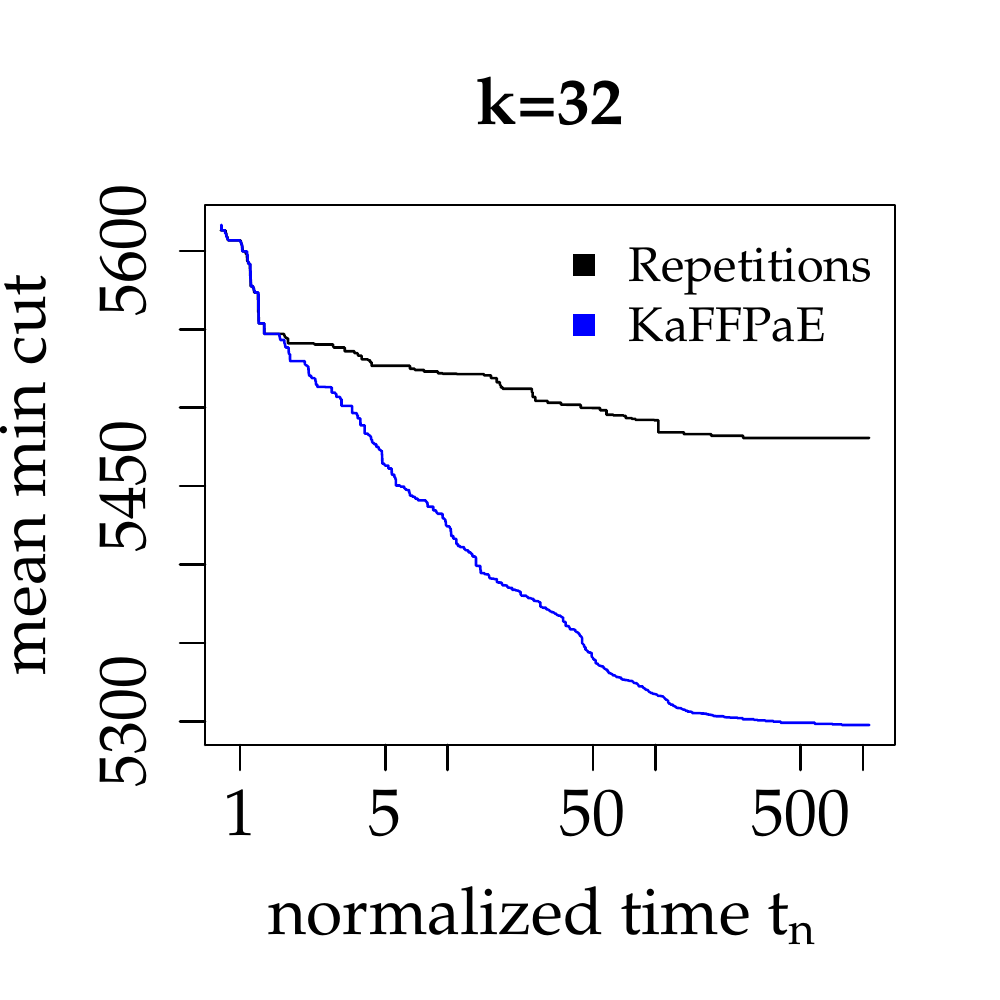}
\includegraphics[width=4cm]{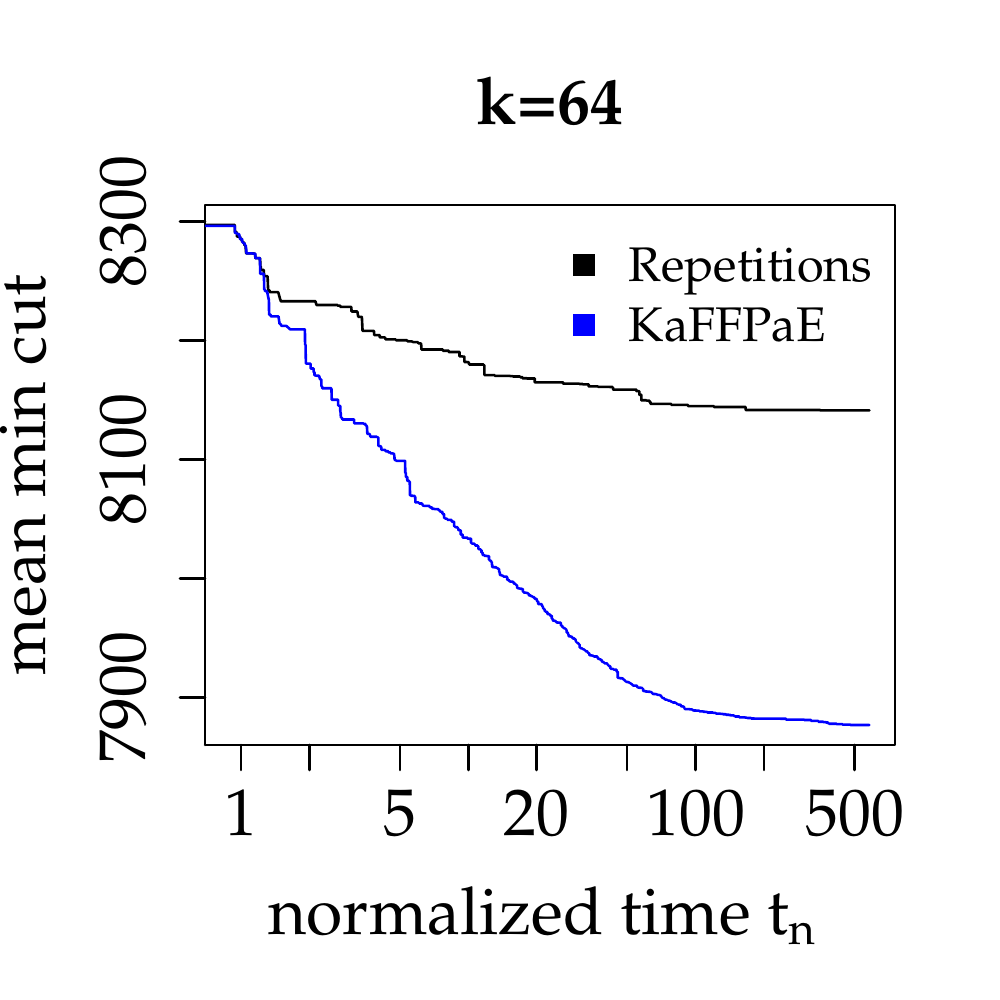}
\includegraphics[width=4cm]{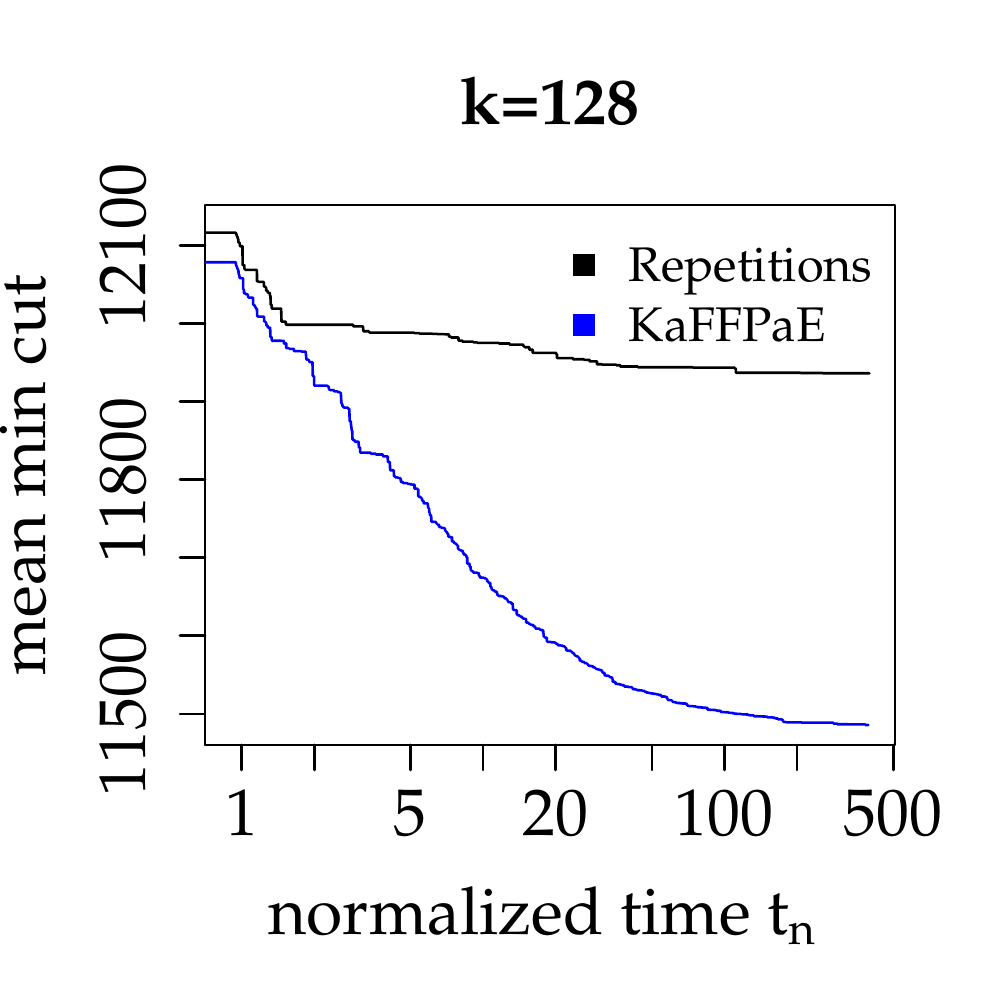}
\includegraphics[width=4cm]{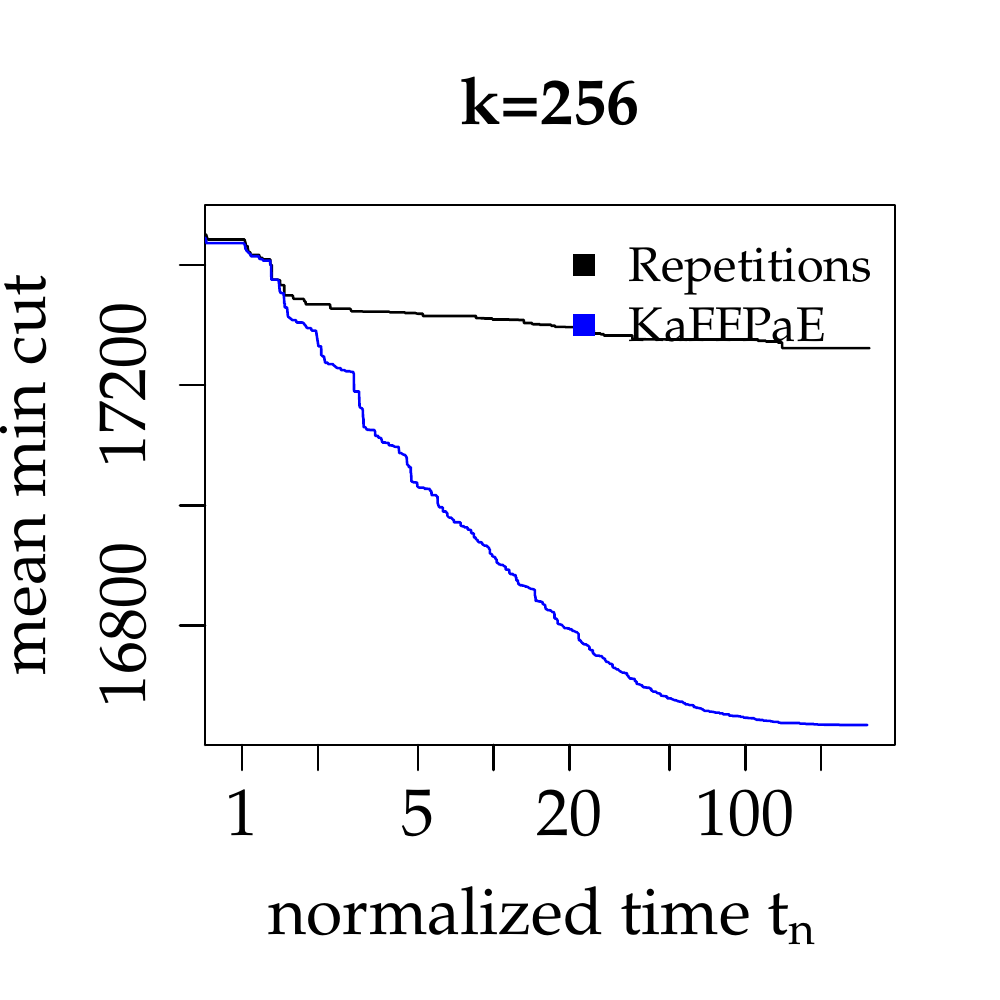}
\end{center}
\vspace*{-0.5cm}
\caption{Convergence plots for the comparison of KaFFPaE with repeated executions of KaFFPa.}
\label{fig:comparision}
\end{figure}

\begin{wraptable}{r}{0.3\textwidth}
\begin{center}
\vspace*{-1cm}
\small
\begin{tabular}{r||r|r}
\hline
$k$/Algo. & Reps. & KaFFPaE \\
        & Avg.  & impr. \%\\
                \hline
                \hline
                2 & \numprint{569}& 0.2\%\\
                4 & \numprint{1229} & 1.0\%\\
                8 &\numprint{2206}& 1.5\%\\
                16 &\numprint{3568}& 2.7\%\\
                32 &\numprint{5481}& 3.4\%\\
                64 &\numprint{8141}& 3.3\%\\
                128 &\numprint{11937}& 3.9\%\\
                256 &\numprint{17262}& 3.7\%\\
                \hline
                \hline
                overall &\numprint{3872}& 2.5\%\\
\hline
\end{tabular}
\end{center}
\vspace*{-0.5cm}

\caption{Different algorithms after two hours of time on 16 PEs.}
\vspace*{-0.25cm}
\end{wraptable}
In this Section we compare ourselves with repeated executions of KaFFPa and other systems.
We switch to our middle sized testset to avoid the effect of overtuning our algorithm parameters to the instances used for calibration. 
We use 16 PEs and two hours of time per instance when we use KaFFPaE.
We parallelized repeated executions of KaFFPa (embarrassingly parallel, different seeds) and also gave 16 PEs and two hours to KaFFPa.
We look at $k \in \{2,4,8,16,32,64,128,256\}$ and performed three repetitions per instance.
Figure~\ref{fig:comparision} show convergence plots for $k \in \{32, 64, 128, 256\}$. All convergence plots can be found in the Appendix~\ref{sec:comparision_all}.
As expected the improvements of KaFFPaE relative to repeated executions of KaFFPa increase with increasing $k$. The largest improvement is obtained for $k=128$. 
Here KaFFPaE produces partitions that have a 3.9\% smaller cut value than plain restarts of the algorithm. 
Note that using a weaker base case partitioner, e.g. KaFFPaEco, increases this value. 
On the small sized testset we obtained an improvement of 5.9\% for $k=64$ compared to plain restarts of KaFFPaEco. 
Tables comparing KaFFPaE with the best results out of ten repetitions of Scotch and Metis can be found in the Appendix Table~\ref{fig:allnumberscomparision}. 
Overall, Scotch and Metis produce 19\% and 28\% larger (best) cuts than KaFFPaE respectively. 
However, these methods are much faster than ours (Appendix Table~\ref{fig:allnumberscomparision}). 

\vspace*{-0.25cm}
\subsection{Combine Operator Experiments}
\label{sec:combineopexperiment}
\begin{wraptable}{l}{0.35\textwidth}
\begin{center}
\vspace*{-1cm}
\small
\begin{tabular}{r||r|r|r|r}
\hline
Algo. & S3R & K3R  & KC & SC \\
                \hline
   $k$     & Avg. & \multicolumn{3}{c}{improvement \%}\\
                \hline
                \hline
                2   & \numprint{591}   & \numprint{2.4} & \numprint{1.6} & \numprint{0.2}   \\
                4   & \numprint{1304}  & \numprint{3.4} & \numprint{4.0} & \numprint{0.2}  \\
                8   & \numprint{2336}  & \numprint{3.7} & \numprint{3.6} & \numprint{0.2}  \\
                16  & \numprint{3723}  & \numprint{2.9} & \numprint{2.0} & \numprint{0.2}  \\
                32  & \numprint{5720}  & \numprint{2.7} & \numprint{3.3} & \numprint{0.0}  \\
                64  & \numprint{8463}  & \numprint{2.8} & \numprint{3.0} & \numprint{-0.6}  \\
                128 & \numprint{12435} & \numprint{3.6} & \numprint{4.5} & \numprint{0.0} \\
                256 & \numprint{17915} & \numprint{3.4} & \numprint{4.2} & \numprint{-0.1} \\
\hline
\end{tabular}
\end{center}

\vspace*{-.5cm}
\caption{Comparison of quality of different algorithms relative to S3R.}
\label{tab:combineexperiementquality}
\vspace*{-.5cm}
\end{wraptable}
We now look into the effectiveness of our combine operator $C_1$.
We conduct the following experiment: we compare the best result of three repeated executions of KaFFPa (\textit{K3R}) against a combine step (\textit{KC}), i.e. after creating two partitions we report the result of the combine step $C_1$ combining both individuals.
The same is done using the combine operator of Soper et. al. \cite{soper2004combined} (\textit{SC}), i.e. we create two individuals using perturbed edge weights as in \cite{soper2004combined} and report the cut produced by the combine step proposed there (the best out of the three individuals). We also present best results out of three repetitions when using perturbed edge weights as in Soper et. al. (\textit{S3R}). 
Since our partitioner does not support double type edge weights, we computed the perturbations and scaled them by a factor of \numprint{10000} (for S3R and SC).
We performed ten repetitions on the middle sized testset. 
Results are reported in Table~\ref{tab:combineexperiementquality}. 
A table presenting absolute average values and comparing the runtime of these algorithms can be found in Appendix Table~\ref{tab:combineexperiementruntime}.
We can see that for large $k$ our new combine operator yields improved partition quality in compareable or less time (KC vs. K3R)). 
Most importantly, we can see that edge biases decrease the solution quality (K3R vs. S3R).
This is due to the fact that edge biases make edge cuts optimial that are not close to optimial in the unbiased problem. 
For example on 2D grid graphs, we have straight edge cuts that are optimal.
Random edge biases make bended edge cuts optimal. 
However, these cuts are are not close to optimal cuts of the original graph partitioning problem.
Moreover, local search algorithms (Flow-based, FM-based) work better if there are a lot of equally sized cuts.

\subsection{Walshaw Benchmark}
\label{sec:walshawbenchmark}
We now apply KaFFPaE to Walshaw's benchmark archive \cite{soper2004combined} using the rules used there, i.e., running time is not an issue but we want to achieve minimal cut values for $k \in \{2,4,8,16,32,64\}$ and balance parameters $\epsilon \in \{0,0.01,0.03,0.05\}$. 
We focus on $\epsilon \in \{1\%,3\%,5\%\}$ since KaFFPaE (more precisely KaFFPa) is not made for the case $\epsilon=0$. 
We run KaFFPaE with a time limit of two hours using 16 PEs (two nodes of the cluster) per graph, $k$ and $\epsilon$ and report the best results obtained in the Appendix~\ref{sec:walshawbenchmarktable}.
KaFFPaE computed 300 partitions which are better than previous best partitions reported there: 91 for 1\%, 103 for 3\% and 106 for 5\%. Moreover, it reproduced \textit{equally sized} cuts in 170 of the 312 remaining cases. 
When only considering the 15 largest graphs and $\epsilon \in \{1.03, 1.05\}$ we are able to reproduce or improve the current result in 224 out of 240 cases. Overall our systems (including KaPPa, KaSPar, KaFFPa, KaFFPaE) now improved or reproduced the entrys in 550 out of 612 cases (for $\epsilon \in \{0.01, 0.03, 0.05\}$).

\vspace*{-.25cm}
\subsection{Comparison with PUNCH}
\label{sec:exproadnetworks}

\begin{wraptable}{r}{0.4\textwidth}
\small
\vspace*{-1cm}
\begin{center}
\begin{tabular}{r||r|r||r|r||r}
\hline
grp, $k$ & \multicolumn{5}{c}{algorithm/runtime $t$}  \\
                \hline
ger.   & P$_{best}$   & $t_{\text{total}}$ & B$_\text{avg}$ & $t_{\text{avg}}$ & B$_\text{best}$ \\
\hline
2      & \numprint{164}  & 83  & 161             & 6  & \textbf{\numprint{161}}    \\
4      & \numprint{400}  & 96  & 394             & 6  & \textbf{\numprint{393}}    \\
8      & \numprint{711}  & 102 & 694             & 9  & \textbf{\numprint{693}}    \\
16     & \numprint{1144} & 83  & \numprint{1148} & 16 & \textbf{\numprint{1137}}   \\
32     & \numprint{1960} & 71  & \numprint{1928} & 31 & \textbf{\numprint{1898}}   \\
64     & \numprint{3165} & 83  & \numprint{3164} & 62 & \textbf{\numprint{3143}}   \\
\hline
\hline
eur. & P$_{best}$ & $t_{\text{total}}$ & B$_\text{avg}$ & $t_{\text{avg}}$ & B$_\text{best}$ \\

                \hline
                2      & \numprint{129}          & 423 & \numprint{149}  & 39  & \textbf{\numprint{129}} \\
                4      & \textbf{\numprint{309}} & 358 & \numprint{313}  & 39  & \numprint{310}          \\
                8      & \textbf{\numprint{634}} & 293 & \numprint{693}  & 47  & \numprint{659}           \\
                16     & \numprint{1293}         & 252 & \numprint{1261} & 73  & \textbf{\numprint{1238}} \\
                32     & \numprint{2289}         & 217 & \numprint{2259} & 130 & \textbf{\numprint{2240}} \\
                64     & \numprint{3828}         & 241 & \numprint{3856} & 248 & \textbf{\numprint{3825}} \\
                \hline
\end{tabular}
\caption{Results on road networks: best results of PUNCH (P) out of 100 repetitions and total time [m] needed to compute these results; average and best cut results of Buffoon (B) as well as average runtime [m] (including preprocessing).}
\vspace*{-0.5cm}
\label{tab:resultsonroadnetworks}
\end{center}
\end{wraptable}
In this Section we focus on finding partitions for road networks. 
We implemented a specialized algorithm, Buffoon, which is similar to PUNCH \cite{delling2010graph} in the sense that it also uses natural cuts as a preprocessing technique to obtain a coarser graph on which the graph partitioning problem is solved. 
For more information on natural cuts, we refer the reader to \cite{delling2010graph}.
Using our (shared memory) parallelized version of natural cut preprocessing we obtain a coarse version of the graph. 
Note that our preprocessing uses slightly different parameters than PUNCH (using the notation of \cite{delling2010graph}, we use $\mathcal{C}=2, U=(1+\epsilon)\frac{n}{2k}, f=10, \alpha=1$).
Since partitions of the coarse graph correspond to partitions of the original graph, we use KaFFPaE to partition the coarse version of the graph. 

After preprocessing, we gave KaFFPaE $t_{\text{eur},k} = k \times 3.75\text{ min}$ on europe and $t_{\text{ger},k} = k \times 0.9375\text{ min}$ on germany, to compute a partition.
In both cases we used all 16 cores (hyperthreading active) of machine B for preprocessing and for KaFFPaE. The experiments where repeated ten times. 
A summary of the results is shown in Table~\ref{tab:resultsonroadnetworks}. 
Interestingly, on germany already our average values are smaller or equal to the best result out of 100 repetitions obtained by PUNCH. 
Overall in 9 out of 12 cases we compute a best cut that is better or equal to the best cut obtained by PUNCH. 
Note that for obtaining the best cut values we invest significantly more time than PUNCH.
However, their machine is about a factor two faster (12 cores running at 3.33GHz compared to 8 cores running at 2.67GHz) and our algorithm is not tuned for road networks.
A table comparing the results on road networks against KaFFPa, KaSPar, Scotch and Metis can be found in Appendix~\ref{tab:detailedroadnetworks}.
These algorithms produce 9\%, 12\%, 93\% and 288\% larger cuts on average respectively.

\vspace*{-.25cm}
\section{Conclusion and Future Work}
KaFFPaE is an distributed evolutionary algorithm to tackle the graph partitioning problem.
Due to new crossover and mutation operators as well as its scalable parallelization it is able to compute the best known partitions for many standard benchmark instances in only a \textit{few minutes}. 
We therefore believe that KaFFPaE is still helpful in the area of high performance computing.

Regarding future work, we want to integrate other partitioners if they implement the possibility to block edges during the coarsening phase and use the given partitioning as initial solution. 
It would be interesting to try other domain specific combine operators, e.g. on social networks it could be interesting to use a modularity clusterer to compute a clustering for the combine operation. 

\bibliographystyle{plain}
\bibliography{quellen}

\begin{thebibliography}{10}

\bibitem{AlbaT02}
Enrique Alba and Marco Tomassini.
\newblock Parallelism and evolutionary algorithms.
\newblock {\em IEEE Trans. Evolutionary Computation}, 6(5):443--462, 2002.

\bibitem{baeckEvoAlgPHD96}
Thomas B\"ack.
\newblock {\em Evolutionary algorithms in theory and practice : evolution
  strategies, evolutionary programming, genetic algorithms}.
\newblock PhD thesis, 1996.

\bibitem{dimacschallengegraphpartandcluster}
David Bader, Henning Meyerhenke, Peter Sanders, and Dorothea Wagner.
\newblock {10th DIMACS Implementation Challenge - Graph Partitioning and Graph
  Clustering, \url{http://www.cc.gatech.edu/dimacs10/}}.

\bibitem{journals/jea/BauerDSSSW10}
Reinhard Bauer, Daniel Delling, Peter Sanders, Dennis Schieferdecker, Dominik
  Schultes, and Dorothea Wagner.
\newblock Combining hierarchical and goal-directed speed-up techniques for
  dijkstra's algorithm.
\newblock {\em ACM Journal of Experimental Algorithmics}, 15, 2010.

\bibitem{conf/ieeeconftoolsartintell/benlichao2010}
Una Benlic and Jin-Kao Hao.
\newblock A multilevel memtetic approach for improving graph $k$-partitions.
\newblock In {\em 22nd Intl. Conf. Tools with Artificial Intelligence}, pages
  121--128, 2010.

\bibitem{boese1994new}
K.D. Boese, A.B. Kahng, and S.~Muddu.
\newblock A new adaptive multi-start technique for combinatorial global
  optimizations.
\newblock {\em Operations Research Letters}, 16(2):101--113, 1994.

\bibitem{journals/ipl/BuiJ92}
Thang~Nguyen Bui and Curt Jones.
\newblock Finding good approximate vertex and edge partitions is {N}{P}-hard.
\newblock {\em Inf. Process. Lett.}, 42(3):153--159, 1992.

\bibitem{journals/tc/ChardaireBM07}
Pierre Chardaire, Musbah Barake, and Geoff~P. McKeown.
\newblock A probe-based heuristic for graph partitioning.
\newblock {\em IEEE Trans. Computers}, 56(12):1707--1720, 2007.

\bibitem{dejongEvoComp2006}
Kenneth~Alan De~Jong.
\newblock {\em Evolutionary computation : a unified approach}.
\newblock MIT Press, 2006.

\bibitem{DSSW09}
D.~Delling, P.~Sanders, D.~Schultes, and D.~Wagner.
\newblock Engineering route planning algorithms.
\newblock In {\em Algorithmics of Large and Complex Networks}, volume 5515 of
  {\em LNCS State-of-the-Art Survey}, pages 117--139. Springer, 2009.

\bibitem{delling2010graph}
Daniel Delling, Andrew~V. Goldberg, Ilya Razenshteyn, and Renato~F. Werneck.
\newblock {Graph Partitioning with Natural Cuts}.
\newblock In {\em 25th International Parallel and Distributed Processing
  Symposium (IPDPS'11)}. IEEE Computer Society, 2011.

\bibitem{conf/icalp/DoerrF11}
Benjamin Doerr and Mahmoud Fouz.
\newblock Asymptotically optimal randomized rumor spreading.
\newblock In {\em ICALP (2)}, volume 6756 of {\em Lecture Notes in Computer
  Science}, pages 502--513. Springer, 2011.

\bibitem{DH03a}
D.~Drake and S.~Hougardy.
\newblock A simple approximation algorithm for the weighted matching problem.
\newblock {\em Information Processing Letters}, 85:211--213, 2003.

\bibitem{fiduccia1982lth}
C.~M. Fiduccia and R.~M. Mattheyses.
\newblock {A Linear-Time Heuristic for Improving Network Partitions}.
\newblock In {\em 19th Conference on Design Automation}, pages 175--181, 1982.

\bibitem{fjallstrom1998agp}
P.O. Fjallstrom.
\newblock {Algorithms for graph partitioning: A survey}.
\newblock {\em Linkoping Electronic Articles in Computer and Information
  Science}, 3(10), 1998.

\bibitem{goldbergGA89}
David~E. Goldberg.
\newblock {\em Genetic algorithms in search, optimization, and machine
  learning}.
\newblock Addison-Wesley, 1989.

\bibitem{kappa}
M.~Holtgrewe, P.~Sanders, and C.~Schulz.
\newblock {Engineering a Scalable High Quality Graph Partitioner}.
\newblock {\em 24th IEEE International Parallal and Distributed Processing
  Symposium}, 2010.

\bibitem{conf/ppsn/InayoshiM94}
Hiroaki Inayoshi and Bernard Manderick.
\newblock The weighted graph bi-partitioning problem: A look at ga performance.
\newblock In {\em PPSN}, volume 866 of {\em Lecture Notes in Computer Science},
  pages 617--625. Springer, 1994.

\bibitem{karypis1999pmk}
G.~Karypis, V.~Kumar, Army High Performance Computing~Research Center, and
  University of~Minnesota.
\newblock {Parallel multilevel k-way partitioning scheme for irregular graphs}.
\newblock {\em SIAM Review}, 41(2):278--300, 1999.

\bibitem{conf/gecco/KimHKM11}
Jin Kim, Inwook Hwang, Yong-Hyuk Kim, and Byung~Ro Moon.
\newblock Genetic approaches for graph partitioning: a survey.
\newblock In {\em GECCO}, pages 473--480. ACM, 2011.

\bibitem{MauSan07}
J.~Maue and P.~Sanders.
\newblock Engineering algorithms for approximate weighted matching.
\newblock In {\em 6th Workshop on Exp. Algorithms ({WEA})}, volume 4525 of {\em
  LNCS}, pages 242--255. Springer, 2007.

\bibitem{Miller95geneticalgorithms}
Brad~L. Miller and David~E. Goldberg.
\newblock Genetic algorithms, tournament selection, and the effects of noise.
\newblock {\em Complex Systems}, 9:193--212, 1995.

\bibitem{kaspar}
V.~Osipov and P.~Sanders.
\newblock {n-Level Graph Partitioning}.
\newblock {\em 18th European Symposium on Algorithms (see also arxiv preprint
  arXiv:1004.4024)}, 2010.

\bibitem{Scotch}
F.~Pellegrini.
\newblock Scotch home page.
\newblock {\url{http://www. labri.fr/pelegrin/scotch}}.

\bibitem{journals/corr/abs-0905-4918}
Josep~M. Pujol, Vijay Erramilli, and Pablo Rodriguez.
\newblock Divide and conquer: Partitioning online social networks.
\newblock {\em CoRR}, abs/0905.4918, 2009.

\bibitem{kaffpa}
P.~Sanders and C.~Schulz.
\newblock {Engineering Multilevel Graph Partitioning Algorithms}.
\newblock {\em 19th European Symposium on Algorithms (see also arxiv preprint
  arXiv:1012.0006v3)}, 2011.

\bibitem{schloegel2000gph}
K.~Schloegel, G.~Karypis, and V.~Kumar.
\newblock {Graph Partitioning for High Performance Scientific Simulations}.
\newblock {\em UMSI research report/University of Minnesota (Minneapolis, Mn).
  Supercomputer institute}, page~38, 2000.

\bibitem{soper2004combined}
A.J. Soper, C.~Walshaw, and M.~Cross.
\newblock A combined evolutionary search and multilevel optimisation approach
  to graph-partitioning.
\newblock {\em Journal of Global Optimization}, 29(2):225--241, 2004.

\bibitem{walshaw2004multilevel}
C.~Walshaw.
\newblock {Multilevel refinement for combinatorial optimisation problems}.
\newblock {\em Annals of Operations Research}, 131(1):325--372, 2004.

\bibitem{walshaw2000mpm}
C.~Walshaw and M.~Cross.
\newblock {Mesh Partitioning: A Multilevel Balancing and Refinement Algorithm}.
\newblock {\em SIAM Journal on Scientific Computing}, 22(1):63--80, 2000.

\bibitem{Walshaw07}
C.~Walshaw and M.~Cross.
\newblock {JOSTLE: Parallel Multilevel Graph-Partitioning Software -- An
  Overview}.
\newblock In F.~Magoules, editor, {\em {Mesh Partitioning Techniques and Domain
  Decomposition Techniques}}, pages 27--58. Civil-Comp Ltd., 2007.
\newblock (Invited chapter).

\end{thebibliography}
\pagebreak
\begin{appendix}
\section{Karlsruhe Fast Flow Partitioner} \label{s:kaffpa}
We now provide a brief overview over the techniques used in the underlying graph partitioner which is used a graph partitioner later. KaFFPa \cite{kaffpa} is a classical matching based multilevel graph partitioner. Recall that a multilevel graph partitioner basically has three phases: coarsening, initial partitioning and uncoarsening. 

KaFFPa makes contraction more systematic by separating two issues: A \emph{rating function} indicates how much sense it makes to contract an edge based on \emph{local} information.  
A \emph{matching} algorithm tries to maximize the sum of the ratings of the contracted edges looking at the \emph{global} structure of the graph. 
While the rating functions allows a flexible characterization of what a ``good'' contracted graph is, the simple, standard definition of the matching problem allows to reuse previously developed algorithms for weighted matching. 
Matchings are contracted until the graph is ``small enough''. 
In \cite{kappa} we have observed that the rating function $\expansion^{*2}(\set{u,v})\Is \frac{\omega(\set{u,v})^2}{c(u)c(v)}$ works best among other edge rating functions, so that this rating function is also used in KaFFPa.

We employed the \emph{Global Path Algorithm (GPA)} as a matching algorithm. 
It was proposed in \cite{MauSan07} as a synthesis of the Greedy algorithm and the Path Growing Algorithm \cite{DH03a}.
This algorithm achieves a half-approximation in the worst case, but empirically, GPA gives considerably better results than Sorted Heavy Edge Matching and Greedy (for more details see \cite{kappa}). 
GPA scans the edges in order of decreasing weight but rather than immediately building a matching, it first constructs a collection of paths and even cycles. 
Afterwards, optimal solutions are computed for each of these paths and cycles using dynamic programming. 

The contraction is stopped when the number of remaining nodes is below $\max{(60k,n/(60k))}$. The graph is then small enough to be partitioned by some initial partitioning algorithm. 
KaFFPa employs Scotch as an initial partitioner since it empirically performs better than Metis. 

Recall that the refinement phase iteratively uncontracts the matchings
contracted during the contraction phase.  After a matching is uncontracted,
local search based refinement algorithms move nodes between block boundaries in
order to reduce the cut while maintaining the balancing constraint. 
Local improvement algorithms are usually variants of the FM-algorithm \cite{fiduccia1982lth}. The algorithm is organized in rounds. In each round, a priority queue $P$ is used which is initialized with all vertices that are incident to more than one block, in a random order. 
The priority is based on the gain $g(v) = \max_P g_P(v)$ where $g_P(v)$ is the decrease in edge cut when moving $v$ to block $P$.  
Ties are broken randomly if there is more than one block that yields the maximum gain when moving $v$ to it. 
Local search then repeatedly looks for the highest gain node $v$. 
Each node is moved at most once within a round. After a node is moved its unmoved neighbors become eligible, i.e. its unmoved neighbors are inserted into the priority queue. 
When a stopping criterion is reached all movements to the best found cut that occurred within the balance constraint are undone.
This process is repeated several times until no improvement is found.

During the uncoarsening phase KaFFPa additionally uses more advanced refinement algorithms. 
The first method is based on  max-flow min-cut computations between pairs of blocks, i.e., a method to improve a given bipartition. 
Roughly speaking, this improvement method is applied between all pairs of blocks that share a non-empty boundary. 
The algorithm basically constructs a flow problem by growing an area around the given boundary vertices of a pair of blocks such that each min cut in this area yields a feasible bipartition of the original graph within the balance constraint. This yields a locally improved $k$-partition of the graph.
The second method for improving a given partition is called multi-try FM. 
Roughly speaking, a $k$-way local search initialized with a \textit{single} boundary node is \textit{repeatedly} started. 
Previous methods are initialized with \textit{all} boundary nodes.

KaFFPa extended the concept of \emph{iterated multilevel algorithms} which was introduced by \cite{walshaw2004multilevel}. The main idea is to iterate the coarsening and uncoarsening phase.
Once the graph is partitioned, edges that are between two blocks are not contracted. 
An F-cycle works as follows: on \emph{each} level we perform at most \emph{two recursive calls} using different random seeds during contraction and local search.  
A second recursive call is only made the second time that the algorithm reaches a particular level. 
As soon as the graph is partitioned, edges that are between blocks are not contracted.  
This ensures nondecreasing quality of the partition since our refinement algorithms guarantee no worsening and break ties randomly. These so called \textit{global search strategies} are more effective than plain restarts of the algorithm.
\section{Additional Experimental Data}
\subsection{Further Parameter Tuning}
\label{sec:furtherparametertuning}
In this Section we perform parameter tuning using KaFFPaEco (a faster but not so powerful as KaFFPaStrong) as a base case partitioner.
We start tuning the fraction parameter $f$. As before we set the flip coin parameter $c$ to one. 
In Figure~\ref{fig:parametertuning} we can see that the algorithm is not too sensitive about the exact choice of this parameter.
As before, larger values of $f$ speed up the convergence rate and improve the result achieved in the end. 
Since $f=50$ is the best parameter in the end, we choose it as our default value.

We now tune the ratio $\frac{c}{10}:\frac{10 - c}{10}$ between mutation to crossover operations. 
For this test we set $f=50$.
The results a similar to the results achieved when using KaFFPaStrong as a base case partitioner. 
Again we can see that for smaller values of $c$ the algorithm is not to sensitive about the exact choice of the parameter. 
When $c=10$, i.e. no crossover operation is performed the convergence speed slows down which yields worse average results in the end.
The results of $c=9$ and $c=1$ are comparable in the end. We choose $c=1$ for consistency.

\begin{figure}[h!]
\vspace*{-1cm}
\begin{center}
\includegraphics[width=0.4\textwidth]{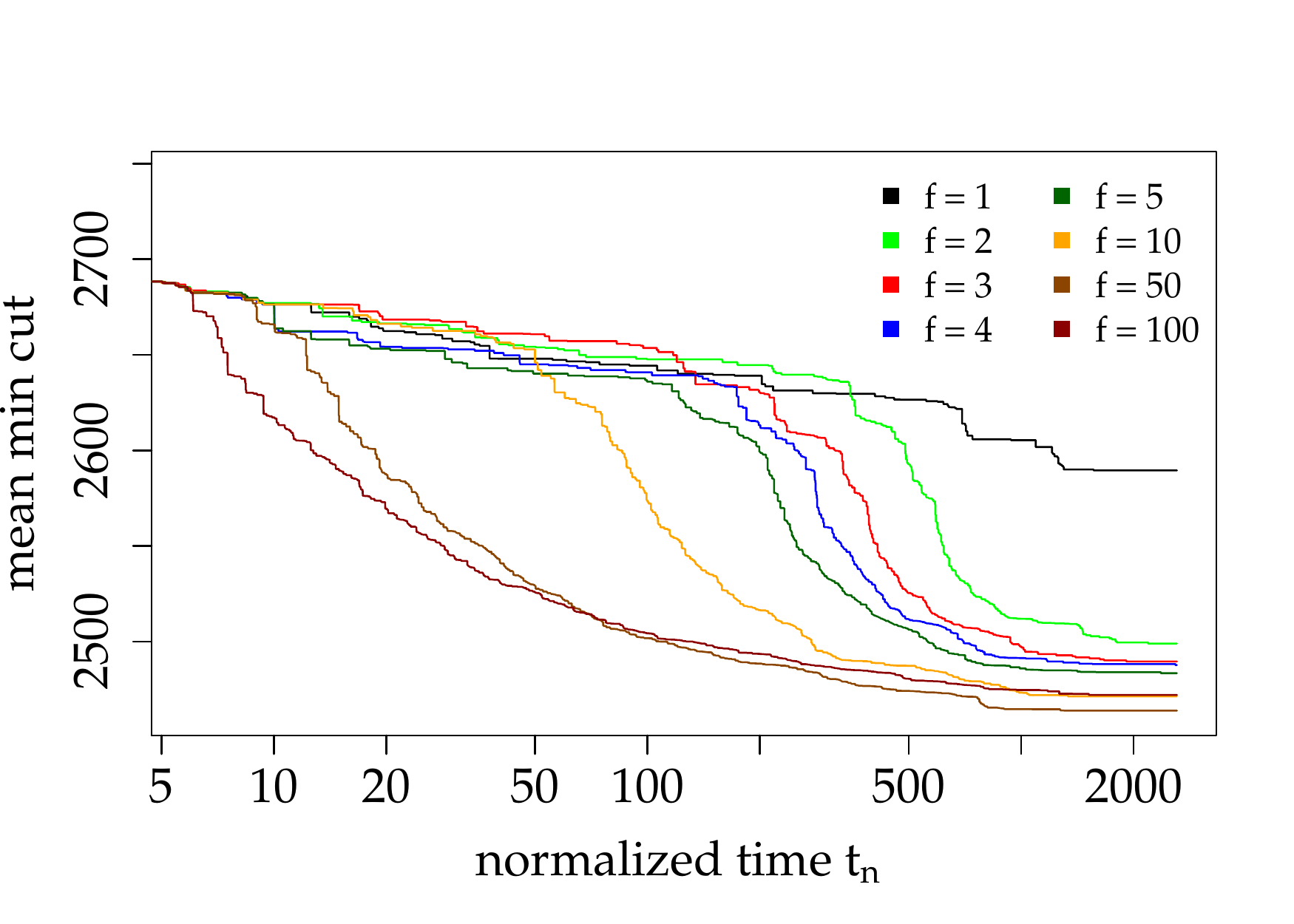}
\includegraphics[width=0.4\textwidth]{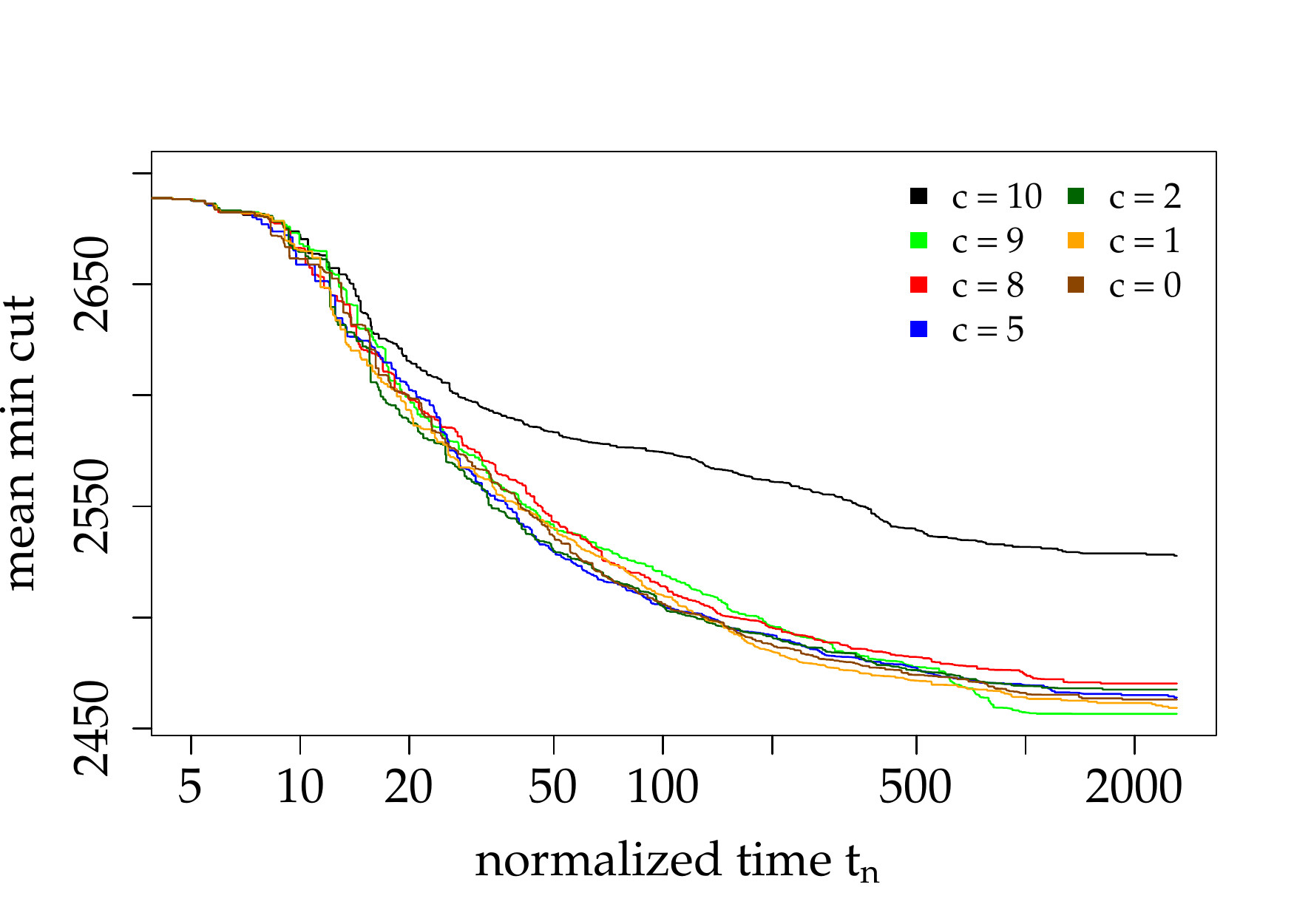}
\vspace*{-.5cm}
\caption{Conv. plots for the \textit{fraction} $f$ using $c=1$ (left) and the \textit{flip coin} $c$ using $f=50$ (right). In both cases KaFFPaEco is used as a base case partitioner. }

\end{center}
\label{fig:parametertuningeco}
\end{figure}

\subsection{Further Comparison Data}
\label{sec:comparision_all}
\begin{figure}[h!]
\begin{center}
\includegraphics[width=4.5cm]{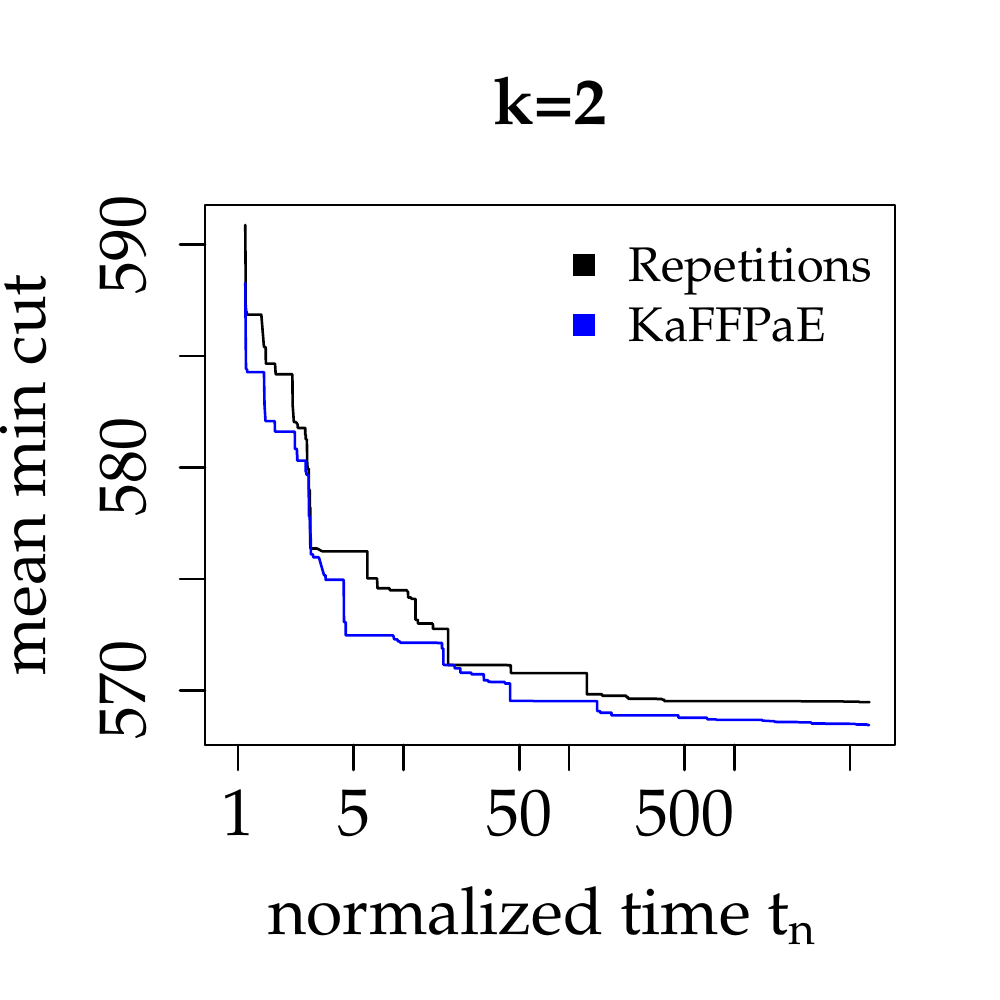}
\includegraphics[width=4.5cm]{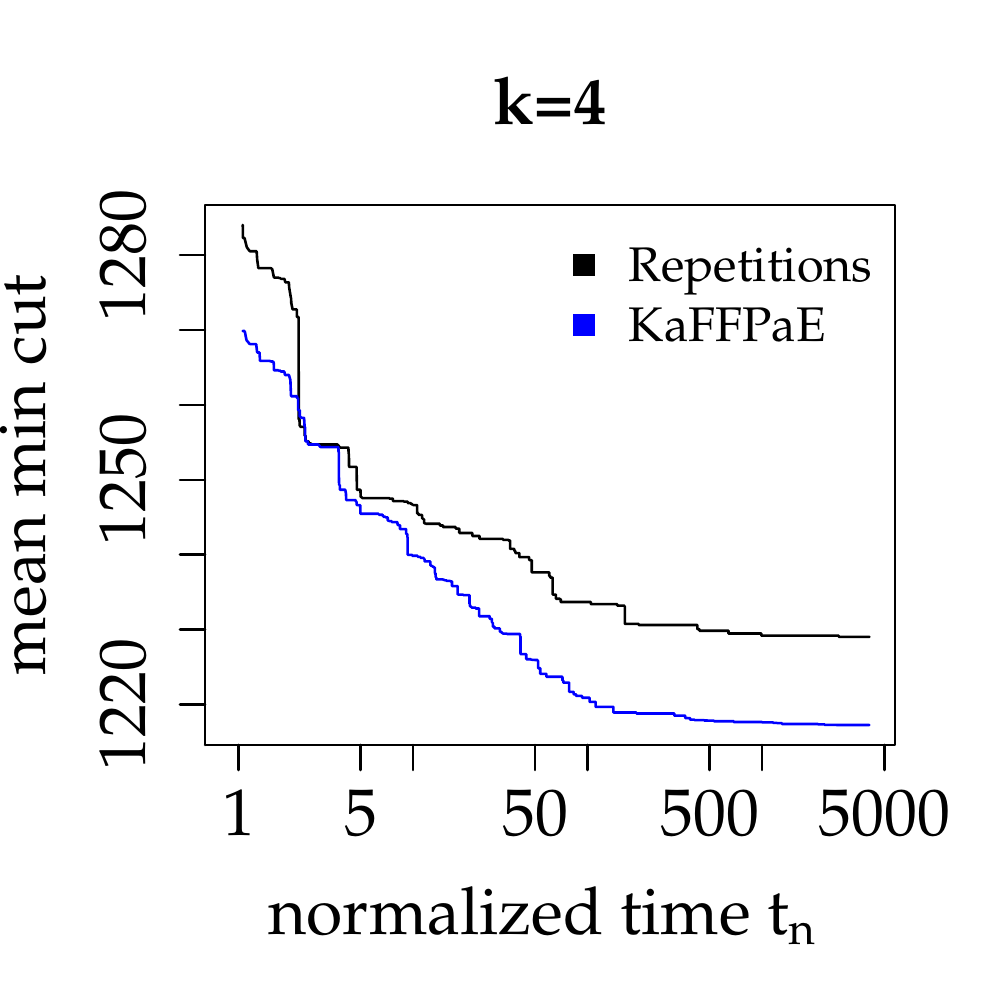} \\
\includegraphics[width=4.5cm]{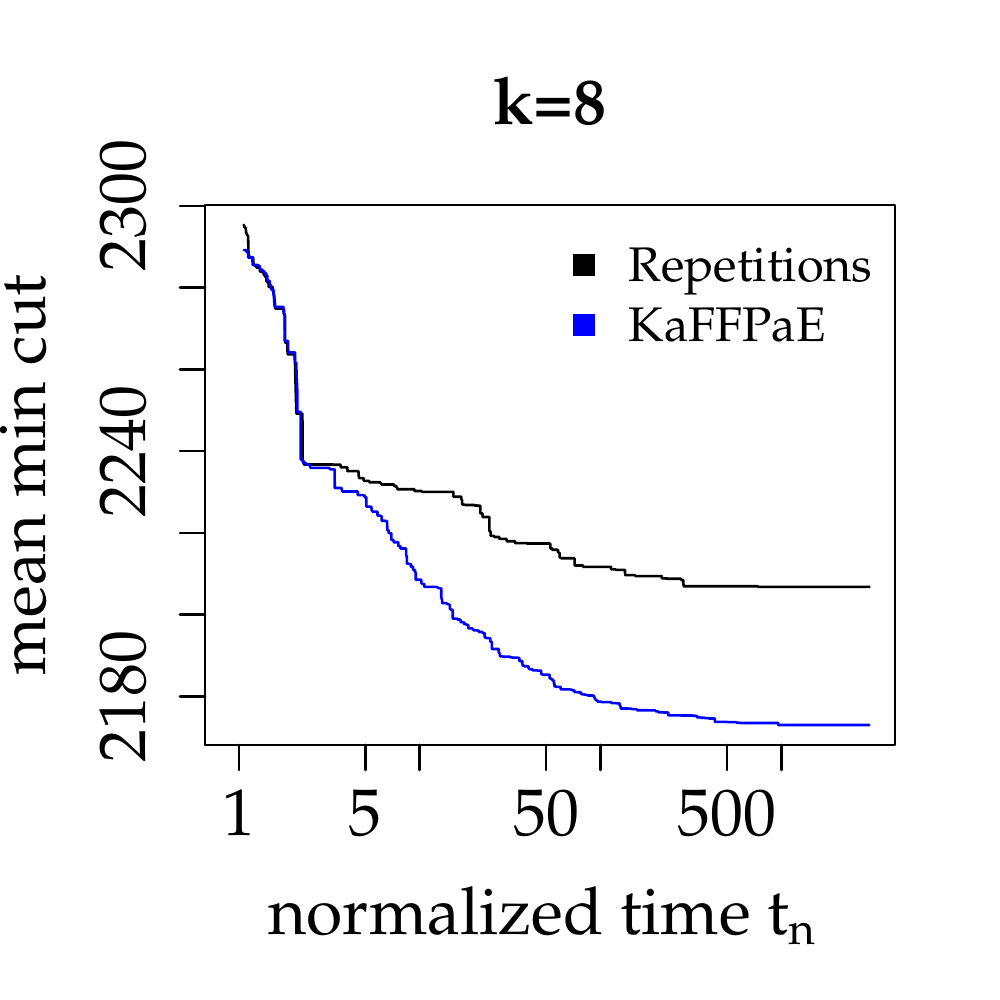} 
\includegraphics[width=4.5cm]{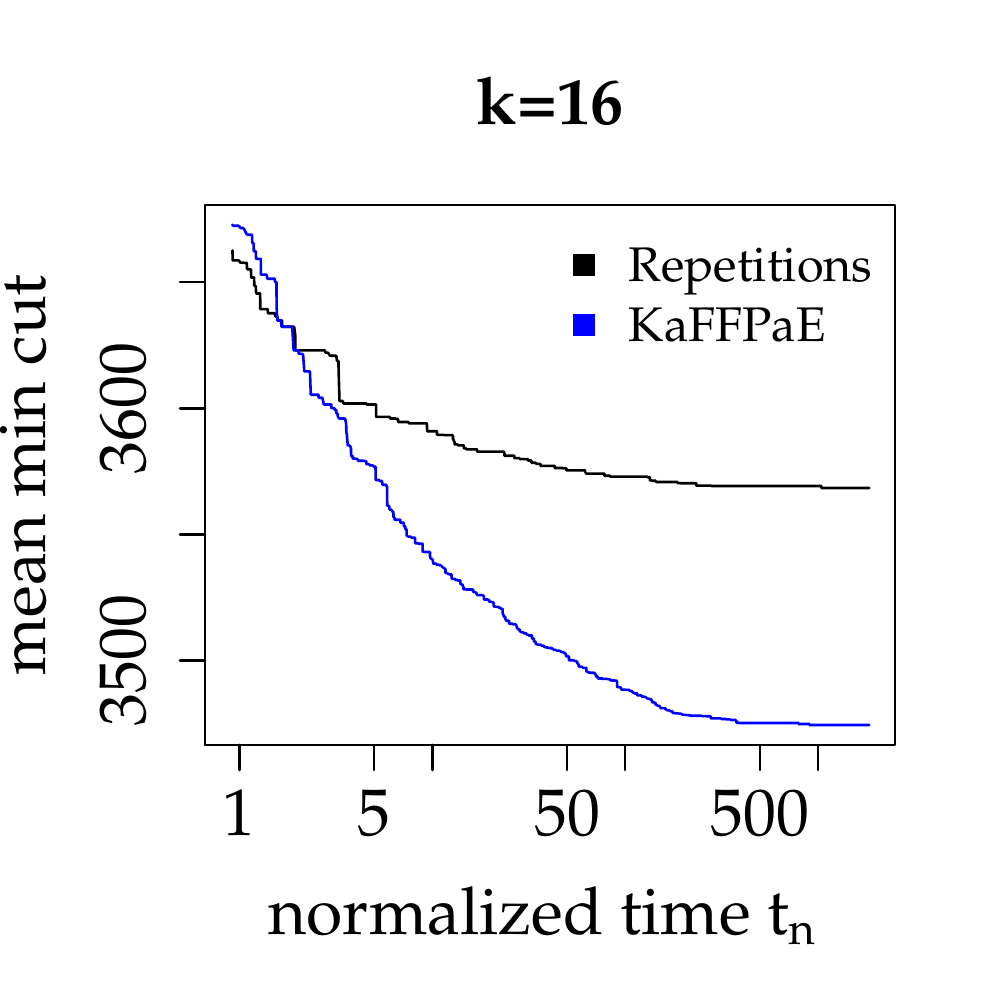} \\
\includegraphics[width=4.5cm]{pics/comparision_middlesize_k32.pdf}
\includegraphics[width=4.5cm]{pics/comparision_middlesize_k64.pdf} \\
\includegraphics[width=4.5cm]{pics/comparision_middlesize_k128.pdf}
\includegraphics[width=4.5cm]{pics/comparision_middlesize_k256.pdf}
\end{center}
\vspace*{-1cm}
\caption{Convergence plots for the comparison with repeated executions of KaFFPa.}
\label{fig:comparision_all}
\end{figure}
\begin{table}[h!]
\begin{center}
\begin{tabular}{r||r|r||r|r||r|r|}
\hline
$k$/Algo. & Reps. & KaFFPaE & \multicolumn{2}{c}{Scotch} & \multicolumn{2}{c|}{Metis} \\
        & Avg. & Avg. & Best. & $t_{\text{avg}}$[s] & Best. & $t_{\text{avg}}$[s] \\
                \hline
                2   & \numprint{569}   & \numprint{568}   &  \numprint{671}   &\numprint{0.22}& \numprint{711}   &\numprint{0.12} \\
                4   & \numprint{1229}  & \numprint{1217}  &  \numprint{1486}  &\numprint{0.41}& \numprint{1574}  &\numprint{0.13} \\
                8   & \numprint{2207}  & \numprint{2173}  &  \numprint{2663}  &\numprint{0.62}& \numprint{2831}  &\numprint{0.13} \\
                16  & \numprint{3568}  & \numprint{3474}  &  \numprint{4192}  &\numprint{0.86}& \numprint{4500}  &\numprint{0.14} \\
                32  & \numprint{5481}  & \numprint{5298}  &  \numprint{6437}  &\numprint{1.15}& \numprint{6899}  &\numprint{0.15} \\
                64  & \numprint{8141}  & \numprint{7879}  &  \numprint{9335}  &\numprint{1.46}& \numprint{10306} &\numprint{0.18} \\
                128 & \numprint{11937} & \numprint{11486} &  \numprint{13427} &\numprint{1.85}& \numprint{14500} &\numprint{0.20} \\
                256 & \numprint{17262} & \numprint{16634} &  \numprint{18972} &\numprint{2.28}& \numprint{20341} &\numprint{0.25} \\
                \hline
                \hline
                overall & \numprint{3872} & \numprint{3779}& \numprint{4507} & 0.87 &\numprint{4835} & 0.16\\
\hline
\end{tabular}
\end{center}
\caption{Averages of final values of different algorithms on the middlesized testset. KaFFPa (Reps) and KaFFPaE was given after two hours of time on 16 PEs per repetitions and instance. Average values of Metis and Scotch are average values of the best cut that occurred out of ten repetitions.}
\label{fig:allnumberscomparision}
\end{table}

\begin{table}
\begin{center}
\vspace*{-1cm}
\small
\begin{tabular}{r||r|r||r|r||r|r||r|r}
\hline
Algo.  & \multicolumn{2}{c}{S3R}                     & \multicolumn{2}{c}{K3R}                                  & \multicolumn{2}{c}{KC} & \multicolumn{2}{c}{SC} \\
\hline                                                                                                   
$k$    & avg.                   & $t$[s]             & avg.                    & $t$[s]                         & avg.             & $t$[s]         & avg.             & $t$[s] \\
\hline                                                                                                   
\hline                                                                                                   
2      & \numprint{591}         & \numprint{19}      & \numprint{577}          & \numprint{14}                  & \numprint{582}   & \numprint{12}  & \numprint{590}   & \numprint{17} \\
4      & \numprint{1304}        & \numprint{30}      & \numprint{1261}         & \numprint{28}                  & \numprint{1254}  & \numprint{22}  & \numprint{1302}  & \numprint{27} \\
8      & \numprint{2336}        & \numprint{40}      & \numprint{2252}         & \numprint{45}                  & \numprint{2255}  & \numprint{36}  & \numprint{2332}  & \numprint{41} \\
16     & \numprint{3723}        & \numprint{54}      & \numprint{3617}         & \numprint{67}                  & \numprint{3649}  & \numprint{57}  & \numprint{3714}  & \numprint{61} \\
32     & \numprint{5720}        & \numprint{82}      & \numprint{5569}         & \numprint{110}                 & \numprint{5540}  & \numprint{99}  & \numprint{5722}  & \numprint{84} \\
64     & \numprint{8463}        & \numprint{116}     & \numprint{8236}         & \numprint{164}                 & \numprint{8213}  & \numprint{146} & \numprint{8512}  & \numprint{113}\\
128    & \numprint{12435}       & \numprint{171}     & \numprint{12008}        & \numprint{239}                 & \numprint{11895} & \numprint{225} & \numprint{12432} & \numprint{162}\\
256    & \numprint{17915}       & \numprint{217}     & \numprint{17335}        & \numprint{327}                 & \numprint{17199} & \numprint{329} & \numprint{17935} & \numprint{232}\\

                \hline
\hline
\end{tabular}
\end{center}

\vspace*{-.5cm}
\caption{Comparison of different combine operators. Average values of cuts and runtime.}
\label{tab:combineexperiementruntime}
\vspace*{-.5cm}
\end{table}
\clearpage
\subsection{Larger Scalability Plots}
\label{sec:largerscalabilityKaFFPaE}
\begin{figure}[h!]
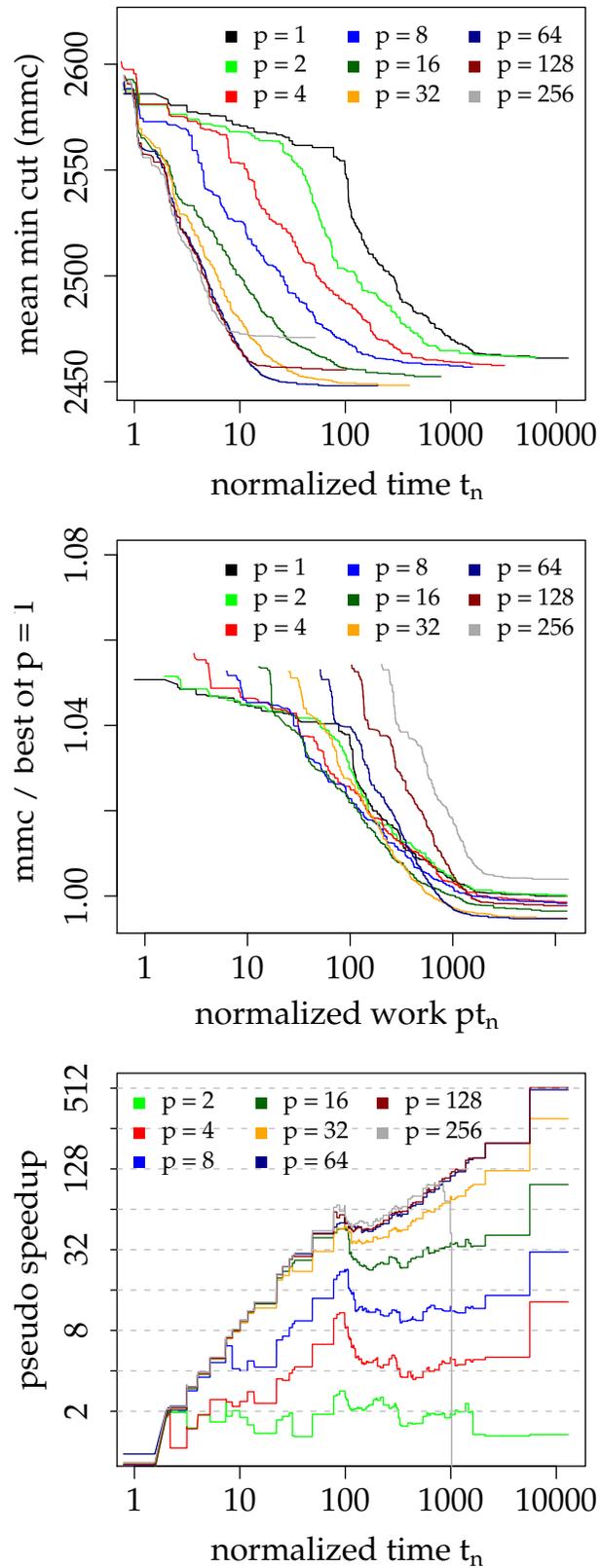

\vspace*{-1.5cm}
\begin{center}
\includegraphics[width=8.5cm]{pics/scalability_stdplot.pdf} \\ 
\vspace*{-1.25cm}
\includegraphics[width=8.5cm]{pics/scalability_nope_normalized_pe1.pdf} \\ 
\vspace*{-1.25cm}
\includegraphics[width=8.5cm]{pics/scalability_speedup.pdf}
\end{center}
\vspace*{-0.5cm}
\caption{Scalability of our algorithm: (upper) a normal convergence plot, (middle) mean minimum cut relative to best cut of KaFFPaE using one PE, (lower) pseudo speedup $S_p(t_n)$.}
\vspace*{-.5cm}       
\end{figure}

\pagebreak
\pagebreak
\clearpage
\subsection{Road Networks}
\begin{table}[H]
\begin{center}
\tiny
\begin{tabular}{l|r|rrr|rrr|rrr|rrr|rrr|rrr|rrr|rrr|}
                &  &   \multicolumn{3}{c|}{PUNCH}        &  \multicolumn{3}{c|}{Buffoon}          & \multicolumn{3}{c|}{KaFFPa Strong}                                                                       & \multicolumn{3}{c|}{KaSPar Strong}                    & \multicolumn{3}{c|}{Scotch} & \multicolumn{3}{c|}{Metis}  \\
graph           & $k$ & Best & Avg. & $t$[m] & Best & Avg. & $t$[m] & Best              & Avg.            & $t$[m]           & Best            & Avg.            & $t$[m]           & Best            & Avg.            & $t$[m]          & Best             & Avg.             & $t$[m]\\
\hline
deu    & 2  & \numprint{164}  & \numprint{166}  & \numprint{0.83} & \numprint{161}  & 161             & \numprint{6.2} & \numprint{163}  & \numprint{166}  & \numprint{3.29}  & \numprint{167}  & \numprint{172}  & \numprint{3.86}  & \numprint{265}  & \numprint{279}  & \numprint{0.05} & \numprint{271}   & \numprint{296}   & \numprint{0.10}\\
deu    & 4  & \numprint{400}  & \numprint{410}  & \numprint{0.96} & \numprint{393}  & 394             & \numprint{6.8} & \numprint{395}  & \numprint{403}  & \numprint{5.25}  & \numprint{419}  & \numprint{426}  & \numprint{4.07}  & \numprint{608}  & \numprint{648}  & \numprint{0.10} & \numprint{592}   & \numprint{710}   & \numprint{0.10}\\
deu    & 8  & \numprint{711}  & \numprint{746}  & \numprint{1.02} & \numprint{693}  & 694             & \numprint{9.7} & \numprint{726}  & \numprint{729}  & \numprint{5.85}  & \numprint{762}  & \numprint{773}  & \numprint{4.17}  & \numprint{1109} & \numprint{1211} & \numprint{0.15} & \numprint{1209}  & \numprint{1600}  & \numprint{0.10}\\
deu    & 16 & \numprint{1144} & \numprint{1188} & \numprint{0.83} & \numprint{1137} & \numprint{1148} & \numprint{16.8} & \numprint{1263} & \numprint{1278} & \numprint{7.05}  & \numprint{1308} & \numprint{1333} & \numprint{4.64}  & \numprint{1957} & \numprint{2061} & \numprint{0.20} & \numprint{2052}  & \numprint{2191}  & \numprint{0.10}\\
deu    & 32 & \numprint{1960} & \numprint{2032} & \numprint{0.71} & \numprint{1898} & \numprint{1928} & \numprint{31.7} & \numprint{2115} & \numprint{2146} & \numprint{7.68}  & \numprint{2182} & \numprint{2217} & \numprint{4.73}  & \numprint{3158} & \numprint{3262} & \numprint{0.25} & \numprint{3225}  & \numprint{3607}  & \numprint{0.10}\\
deu    & 64 & \numprint{3165} & \numprint{3253} & \numprint{0.83} & \numprint{3143} & \numprint{3164} & \numprint{61.1} & \numprint{3432} & \numprint{3440} & \numprint{8.55}  & \numprint{3610} & \numprint{3631} & \numprint{4.89}  & \numprint{4799} & \numprint{4937} & \numprint{0.30} & \numprint{4985}  & \numprint{5320}  & \numprint{0.10}\\
eur    & 2  & \numprint{129}  & \numprint{130}  & \numprint{4.25} & \numprint{129}  & \numprint{175}  & \numprint{39.5} & \numprint{130}  & \numprint{130}  & \numprint{16.88} & \numprint{133}  & \numprint{138}  & \numprint{32.44} & \numprint{369}  & \numprint{448}  & \numprint{0.20} & \numprint{412}   & \numprint{454}   & \numprint{0.55} \\
eur    & 4  & \numprint{309}  & \numprint{309}  & \numprint{3.58} & \numprint{310}  & \numprint{317}  & \numprint{39.1} & \numprint{412}  & \numprint{430}  & \numprint{30.40} & \numprint{355}  & \numprint{375}  & \numprint{36.13} & \numprint{727}  & \numprint{851}  & \numprint{0.40} & \numprint{902}   & \numprint{1698}  & \numprint{0.54} \\
eur    & 8  & \numprint{634}  & \numprint{671}  & \numprint{2.93} & \numprint{659}  & \numprint{671}  & \numprint{47.9} & \numprint{749}  & \numprint{772}  & \numprint{34.45} & \numprint{774}  & \numprint{786}  & \numprint{37.21} & \numprint{1338} & \numprint{1461} & \numprint{0.60} & \numprint{2473}  & \numprint{3819}  & \numprint{0.55} \\
eur    & 16 & \numprint{1293} & \numprint{1353} & \numprint{2.52} & \numprint{1238} & \numprint{1257} & \numprint{73.5} & \numprint{1454} & \numprint{1493} & \numprint{39.01} & \numprint{1401} & \numprint{1440} & \numprint{42.56} & \numprint{2478} & \numprint{2563} & \numprint{0.81} & \numprint{3314}  & \numprint{8554}  & \numprint{0.56} \\
eur    & 32 & \numprint{2289} & \numprint{2362} & \numprint{2.17} & \numprint{2240} & \numprint{2260} & \numprint{130.2} & \numprint{2428} & \numprint{2504} & \numprint{40.76} & \numprint{2595} & \numprint{2643} & \numprint{43.31} & \numprint{4057} & \numprint{4249} & \numprint{1.00} & \numprint{5811}  & \numprint{7380}  & \numprint{0.55} \\
eur    & 64 & \numprint{3828} & \numprint{3984} & \numprint{2.41} & \numprint{3825} & \numprint{3862} & \numprint{248.9} & \numprint{4240} & \numprint{4264} & \numprint{42.23} & \numprint{4502} & \numprint{4526} & \numprint{42.23} & \numprint{6518} & \numprint{6739} & \numprint{1.23} & \numprint{10264} & \numprint{13947} & \numprint{0.55} \\
\hline
\hline
overall &     & \numprint{822}     & \numprint{847}    & \numprint{1.57}  &   \numprint{812}   & \numprint{831}     & \numprint{33.9}  & \numprint{893.05} & \numprint{909}  & \numprint{13.97} & \numprint{911}  & \numprint{931}  & \numprint{13.03} & \numprint{1495} & \numprint{1607} & \numprint{0.30} & \numprint{1800}  & \numprint{2400}  & \numprint{0.23} \\
\hline
\end{tabular}\\
\end{center}
\caption{Detailed per instance results for road networks. PUNCH was run 100 times, Buffoon 10 times and KaFFPa, KaSPar, Scotch and Metis where run 5 times.}
\label{tab:detailedroadnetworks}
\end{table}

\section{Instances}
\label{sec:instances}
\begin{table}[h!]
\begin{center}
        
\begin{tabular}{|l|r|r|}
\hline
\multicolumn{3}{|c|}{small sized instances} \\
\hline
graph & $n$ & $m$\\
\hline
\hline
rgg15 & $2^{15}$  & \numprint{160240} \\
rgg16 &  $2^{16}$ & \numprint{342127} \\
\hline
\hline
delaunay15 & $2^{15}$ &\numprint{98274} \\
delaunay16 & $2^{16}$ &\numprint{196575} \\
\hline
\hline
uk & \numprint{4824}& \numprint{6837}\\
luxemburg& \numprint{114599} & \numprint{119666} \\
\hline
\hline
3elt& \numprint{4720} & \numprint{13722} \\
4elt& \numprint{15606} & \numprint{45878} \\
fe\_sphere& \numprint{16386} & \numprint{49152} \\
cti& \numprint{16840} & \numprint{48232} \\
fe\_body& \numprint{45087} & \numprint{163734} \\
\hline
\end{tabular}
\begin{tabular}{|l|r|r|}
\hline
\multicolumn{3}{|c|}{medium sized instances} \\
\hline
graph & $n$ & $m$\\
\hline
\hline
rgg17 & $2^{17}$  & \numprint{728753} \\
rgg18 &  $2^{18}$ & \numprint{1547283} \\
\hline
\hline
delaunay17 & $2^{17}$ &\numprint{393176} \\
delaunay18 & $2^{18}$ &\numprint{786396} \\
\hline
\hline
bel & \numprint{463514}& \numprint{591882}\\
nld & \numprint{893041}& \numprint{1139540}\\
\hline
\hline
t60k & \numprint{60005} & \numprint{89440} \\
wing & \numprint{62032} & \numprint{121544} \\
fe\_tooth & \numprint{78136} & \numprint{452591} \\
fe\_rotor & \numprint{99617} & \numprint{662431} \\
\hline
\hline
memplus& \numprint{17758} & \numprint{54196}  \\
\hline
\end{tabular}

\begin{tabular}{|l|r|r|}

\hline
\multicolumn{3}{|c|}{road networks} \\
\hline
graph & $n$ & $m$\\
\hline
\hline
germany & \numprint{4378446} & \numprint{5483587}  \\
europe & \numprint{18029721} & \numprint{22217686} \\
\hline
\end{tabular}

\end{center}

\caption{Basic properties of our benchmark set.}
\label{tab:instances}
\label{tab:socialnetworksproperties}
\end{table}

\section{Detailed Walshaw Benchmark Results}
\label{sec:walshawbenchmarktable}
\small
\begin{landscape}

\begin{table}[H]
\small
\begin{center}
\begin{tabular}{|l||r|r||r|r||r|r||r|r||r|r||r|r|}\hline
\small

Graph/$k$  & \multicolumn{2}{|c|}{2} & \multicolumn{2}{|c|}{4} & \multicolumn{2}{|c|}{8} & \multicolumn{2}{|c|}{16} & \multicolumn{2}{|c|}{32} & \multicolumn{2}{|c|}{64}\\

        \hline 
        add20 & \numprint{642} & \textbf{\numprint{594}} &\numprint{1194} & \textbf{\numprint{1159}} &\numprint{1727} & \textbf{\numprint{1696}} &\numprint{2107} & \textbf{\numprint{2062}} &\textbf{\numprint{2512}} & \numprint{2687} &\numprint{3188} & \textbf{\numprint{3108}} \\ 
        data & \textbf{\numprint{188}} & \numprint{188} &\textbf{\numprint{377}} & \numprint{378} &\textbf{\numprint{656}} & \numprint{659} &\numprint{1142} & \textbf{\numprint{1135}} &\numprint{1933} & \textbf{\numprint{1858}} &\numprint{2966} & \textbf{\numprint{2885}} \\ 
        3elt & \textbf{\numprint{89}} & \numprint{89} &\textbf{\numprint{199}} & \numprint{199} &\textbf{\numprint{340}} & \numprint{341} &\textbf{\numprint{568}} & \numprint{569} &\textbf{\numprint{967}} & \numprint{968} &\textbf{\numprint{1553}} & \numprint{1553} \\ 
        uk & \textbf{\numprint{19}} & \numprint{19} &\textbf{\numprint{40}} & \numprint{40} &\textbf{\numprint{80}} & \numprint{82} &\textbf{\numprint{144}} & \numprint{146} &\textbf{\numprint{251}} & \numprint{256} &\textbf{\numprint{417}} & \numprint{419} \\ 
        add32 & \textbf{\numprint{10}} & \numprint{10} &\textbf{\numprint{33}} & \numprint{33} &\textbf{\numprint{66}} & \numprint{66} &\textbf{\numprint{117}} & \numprint{117} &\textbf{\numprint{212}} & \numprint{212} &\textbf{\numprint{486}} & \numprint{493} \\ 
        bcsstk33 & \textbf{\numprint{10096}} & \numprint{10097} &\textbf{\numprint{21390}} & \numprint{21508} &\textbf{\numprint{34174}} & \numprint{34178} &\numprint{55327} & \textbf{\numprint{54763}} &\numprint{78199} & \textbf{\numprint{77964}} &\numprint{109811} & \textbf{\numprint{108467}} \\ 
        whitaker3 & \textbf{\numprint{126}} & \numprint{126} &\textbf{\numprint{380}} & \numprint{380} &\textbf{\numprint{654}} & \numprint{655} &\textbf{\numprint{1091}} & \numprint{1091} &\textbf{\numprint{1678}} & \numprint{1697} &\textbf{\numprint{2532}} & \numprint{2552} \\ 
        crack & \textbf{\numprint{183}} & \numprint{183} &\textbf{\numprint{362}} & \numprint{362} &\textbf{\numprint{676}} & \numprint{677} &\numprint{1098} & \textbf{\numprint{1089}} &\numprint{1697} & \textbf{\numprint{1687}} &\numprint{2581} & \textbf{\numprint{2555}} \\ 
        wing\_nodal & \textbf{\numprint{1695}} & \numprint{1695} &\textbf{\numprint{3563}} & \numprint{3565} &\textbf{\numprint{5422}} & \numprint{5427} &\numprint{8353} & \textbf{\numprint{8339}} &\numprint{12040} & \textbf{\numprint{11828}} &\numprint{16185} & \textbf{\numprint{16124}} \\ 
        fe\_4elt2 & \textbf{\numprint{130}} & \numprint{130} &\textbf{\numprint{349}} & \numprint{349} &\textbf{\numprint{603}} & \numprint{604} &\textbf{\numprint{1002}} & \numprint{1005} &\textbf{\numprint{1620}} & \numprint{1628} &\numprint{2530} & \textbf{\numprint{2519}} \\ 
        vibrobox & \numprint{11538} & \textbf{\numprint{10310}} &\textbf{\numprint{18956}} & \numprint{19098} &\textbf{\numprint{24422}} & \numprint{24509} &\numprint{33501} & \textbf{\numprint{32102}} &\numprint{41725} & \textbf{\numprint{40085}} &\numprint{49012} & \textbf{\numprint{47651}} \\ 
        bcsstk29 & \textbf{\numprint{2818}} & \numprint{2818} &\textbf{\numprint{8029}} & \numprint{8029} &\textbf{\numprint{13904}} & \numprint{13950} &\numprint{22618} & \textbf{\numprint{21768}} &\numprint{35654} & \textbf{\numprint{34841}} &\numprint{57712} & \textbf{\numprint{57031}} \\ 
        4elt & \textbf{\numprint{138}} & \numprint{138} &\textbf{\numprint{320}} & \numprint{320} &\textbf{\numprint{532}} & \numprint{533} &\textbf{\numprint{932}} & \numprint{934} &\numprint{1551} & \textbf{\numprint{1547}} &\textbf{\numprint{2574}} & \numprint{2579} \\ 
        fe\_sphere & \textbf{\numprint{386}} & \numprint{386} &\textbf{\numprint{766}} & \numprint{766} &\textbf{\numprint{1152}} & \numprint{1152} &\textbf{\numprint{1709}} & \numprint{1709} &\numprint{2494} & \textbf{\numprint{2488}} &\numprint{3599} & \textbf{\numprint{3584}} \\ 
        cti & \textbf{\numprint{318}} & \numprint{318} &\textbf{\numprint{944}} & \numprint{944} &\textbf{\numprint{1749}} & \numprint{1752} &\textbf{\numprint{2804}} & \numprint{2837} &\textbf{\numprint{4117}} & \numprint{4129} &\numprint{5820} & \textbf{\numprint{5818}} \\ 
        memplus & \numprint{5491} & \textbf{\numprint{5484}} &\textbf{\numprint{9448}} & \numprint{9500} &\numprint{11807} & \textbf{\numprint{11776}} &\numprint{13250} & \textbf{\numprint{13001}} &\numprint{15187} & \textbf{\numprint{14107}} &\numprint{17183} & \textbf{\numprint{16543}} \\ 
        cs4 & \textbf{\numprint{366}} & \numprint{366} &\textbf{\numprint{925}} & \numprint{934} &\textbf{\numprint{1436}} & \numprint{1448} &\textbf{\numprint{2087}} & \numprint{2105} &\textbf{\numprint{2910}} & \numprint{2938} &\textbf{\numprint{4032}} & \numprint{4051} \\ 
        bcsstk30 & \textbf{\numprint{6335}} & \numprint{6335} &\textbf{\numprint{16596}} & \numprint{16622} &\textbf{\numprint{34577}} & \numprint{34604} &\numprint{70945} & \textbf{\numprint{70604}} &\numprint{116128} & \textbf{\numprint{113788}} &\numprint{176099} & \textbf{\numprint{172929}} \\ 
        bcsstk31 & \textbf{\numprint{2699}} & \numprint{2699} &\textbf{\numprint{7282}} & \numprint{7287} &\textbf{\numprint{13201}} & \numprint{13230} &\textbf{\numprint{23761}} & \numprint{23807} &\numprint{37995} & \textbf{\numprint{37652}} &\numprint{59318} & \textbf{\numprint{58076}} \\ 
        fe\_pwt & \textbf{\numprint{340}} & \numprint{340} &\textbf{\numprint{704}} & \numprint{704} &\textbf{\numprint{1433}} & \numprint{1437} &\textbf{\numprint{2797}} & \numprint{2798} &\textbf{\numprint{5523}} & \numprint{5549} &\textbf{\numprint{8222}} & \numprint{8276} \\ 
        bcsstk32 & \textbf{\numprint{4667}} & \numprint{4667} &\textbf{\numprint{9195}} & \numprint{9208} &\textbf{\numprint{20204}} & \numprint{20323} &\textbf{\numprint{35936}} & \numprint{36399} &\numprint{61533} & \textbf{\numprint{60776}} &\numprint{94523} & \textbf{\numprint{91863}} \\ 
        fe\_body & \textbf{\numprint{262}} & \numprint{262} &\textbf{\numprint{598}} & \numprint{598} &\textbf{\numprint{1026}} & \numprint{1048} &\textbf{\numprint{1714}} & \numprint{1779} &\textbf{\numprint{2796}} & \numprint{2935} &\textbf{\numprint{4825}} & \numprint{4879} \\ 
        t60k & \textbf{\numprint{75}} & \numprint{75} &\textbf{\numprint{208}} & \numprint{208} &\textbf{\numprint{454}} & \numprint{454} &\textbf{\numprint{805}} & \numprint{815} &\textbf{\numprint{1320}} & \numprint{1352} &\textbf{\numprint{2079}} & \numprint{2123} \\ 
        wing & \textbf{\numprint{784}} & \numprint{784} &\textbf{\numprint{1610}} & \numprint{1613} &\textbf{\numprint{2479}} & \numprint{2505} &\textbf{\numprint{3857}} & \numprint{3880} &\textbf{\numprint{5584}} & \numprint{5626} &\numprint{7680} & \textbf{\numprint{7656}} \\ 
        brack2 & \textbf{\numprint{708}} & \numprint{708} &\textbf{\numprint{3013}} & \numprint{3013} &\textbf{\numprint{7040}} & \numprint{7099} &\textbf{\numprint{11636}} & \numprint{11649} &\numprint{17508} & \textbf{\numprint{17398}} &\numprint{26226} & \textbf{\numprint{25913}} \\ 
        finan512 & \textbf{\numprint{162}} & \numprint{162} &\textbf{\numprint{324}} & \numprint{324} &\textbf{\numprint{648}} & \numprint{648} &\textbf{\numprint{1296}} & \numprint{1296} &\textbf{\numprint{2592}} & \numprint{2592} &\textbf{\numprint{10560}} & \numprint{10560} \\ 
        \hline
        fe\_tooth & \textbf{\numprint{3814}} & \numprint{3815} &\textbf{\numprint{6846}} & \numprint{6867} &\textbf{\numprint{11408}} & \numprint{11473} &\numprint{17411} & \textbf{\numprint{17396}} &\numprint{25111} & \textbf{\numprint{24933}} &\numprint{34824} & \textbf{\numprint{34433}} \\ 
        fe\_rotor & \textbf{\numprint{2031}} & \numprint{2031} &\textbf{\numprint{7180}} & \numprint{7292} &\textbf{\numprint{12726}} & \numprint{12813} &\numprint{20555} & \textbf{\numprint{20438}} &\numprint{31428} & \textbf{\numprint{31233}} &\numprint{46372} & \textbf{\numprint{45911}} \\ 
        598a & \textbf{\numprint{2388}} & \numprint{2388} &\textbf{\numprint{7948}} & \numprint{7952} &\numprint{15956} & \textbf{\numprint{15924}} &\textbf{\numprint{25741}} & \numprint{25789} &\numprint{39423} & \textbf{\numprint{38627}} &\numprint{57497} & \textbf{\numprint{56179}} \\ 
        fe\_ocean & \textbf{\numprint{387}} & \numprint{387} &\textbf{\numprint{1816}} & \numprint{1824} &\textbf{\numprint{4091}} & \numprint{4134} &\numprint{7846} & \textbf{\numprint{7771}} &\textbf{\numprint{12711}} & \numprint{12811} &\numprint{20301} & \textbf{\numprint{19989}} \\ 
        144 & \textbf{\numprint{6478}} & \numprint{6478} &\numprint{15152} & \textbf{\numprint{15140}} &\textbf{\numprint{25273}} & \numprint{25279} &\textbf{\numprint{37896}} & \numprint{38212} &\textbf{\numprint{56550}} & \numprint{56868} &\textbf{\numprint{79198}} & \numprint{80406} \\ 
        wave & \textbf{\numprint{8658}} & \numprint{8665} &\textbf{\numprint{16780}} & \numprint{16875} &\textbf{\numprint{28979}} & \numprint{29115} &\textbf{\numprint{42516}} & \numprint{42929} &\textbf{\numprint{61104}} & \numprint{62551} &\textbf{\numprint{85589}} & \numprint{86086} \\ 
        m14b & \textbf{\numprint{3826}} & \numprint{3826} &\textbf{\numprint{12973}} & \numprint{12981} &\textbf{\numprint{25690}} & \numprint{25852} &\numprint{42523} & \textbf{\numprint{42351}} &\textbf{\numprint{65835}} & \numprint{67423} &\textbf{\numprint{98211}} & \numprint{99655} \\ 
        auto & \textbf{\numprint{9949}} & \numprint{9954} &\textbf{\numprint{26614}} & \numprint{26649} &\numprint{45557} & \textbf{\numprint{45470}} &\numprint{77097} & \textbf{\numprint{77005}} &\textbf{\numprint{121032}} & \numprint{121608} &\textbf{\numprint{172167}} & \numprint{174482} \\ 
        \hline
        \end{tabular}
        \end{center} \caption{Computing partitions from scratch $\epsilon = 1$\%. In each $k$-column the results computed by KaFFPaE are on the left and the current Walshaw cuts are presented on the right side. }
        \end{table}
        \end{landscape}

\small
\begin{landscape}

\begin{table}[H]
\small
\begin{center}
\begin{tabular}{|l||r|r||r|r||r|r||r|r||r|r||r|r|}\hline
\small

Graph/$k$  & \multicolumn{2}{|c|}{2} & \multicolumn{2}{|c|}{4} & \multicolumn{2}{|c|}{8} & \multicolumn{2}{|c|}{16} & \multicolumn{2}{|c|}{32} & \multicolumn{2}{|c|}{64}\\

        \hline 
        add20 & \numprint{623} & \textbf{\numprint{576}} &\numprint{1180} & \textbf{\numprint{1158}} &\numprint{1696} & \textbf{\numprint{1689}} &\numprint{2075} & \textbf{\numprint{2062}} &\numprint{2422} & \textbf{\numprint{2387}} &\textbf{\numprint{2963}} & \numprint{3021} \\ 
        data & \textbf{\numprint{185}} & \numprint{185} &\textbf{\numprint{369}} & \numprint{369} &\textbf{\numprint{638}} & \numprint{638} &\textbf{\numprint{1111}} & \numprint{1118} &\numprint{1815} & \textbf{\numprint{1801}} &\numprint{2905} & \textbf{\numprint{2809}} \\ 
        3elt & \textbf{\numprint{87}} & \numprint{87} &\textbf{\numprint{198}} & \numprint{198} &\textbf{\numprint{334}} & \numprint{335} &\textbf{\numprint{561}} & \numprint{562} &\textbf{\numprint{950}} & \numprint{950} &\numprint{1537} & \textbf{\numprint{1532}} \\ 
        uk & \textbf{\numprint{18}} & \numprint{18} &\textbf{\numprint{39}} & \numprint{39} &\textbf{\numprint{78}} & \numprint{78} &\textbf{\numprint{140}} & \numprint{141} &\textbf{\numprint{240}} & \numprint{245} &\textbf{\numprint{406}} & \numprint{411} \\ 
        add32 & \textbf{\numprint{10}} & \numprint{10} &\textbf{\numprint{33}} & \numprint{33} &\textbf{\numprint{66}} & \numprint{66} &\textbf{\numprint{117}} & \numprint{117} &\textbf{\numprint{212}} & \numprint{212} &\textbf{\numprint{486}} & \numprint{490} \\ 
        bcsstk33 & \textbf{\numprint{10064}} & \numprint{10064} &\textbf{\numprint{20767}} & \numprint{20854} &\textbf{\numprint{34068}} & \numprint{34078} &\numprint{54772} & \textbf{\numprint{54455}} &\numprint{77549} & \textbf{\numprint{77353}} &\numprint{108645} & \textbf{\numprint{107011}} \\ 
        whitaker3 & \textbf{\numprint{126}} & \numprint{126} &\textbf{\numprint{378}} & \numprint{378} &\textbf{\numprint{650}} & \numprint{651} &\textbf{\numprint{1084}} & \numprint{1086} &\textbf{\numprint{1662}} & \numprint{1673} &\textbf{\numprint{2498}} & \numprint{2499} \\ 
        crack & \textbf{\numprint{182}} & \numprint{182} &\textbf{\numprint{360}} & \numprint{360} &\textbf{\numprint{671}} & \numprint{673} &\textbf{\numprint{1077}} & \numprint{1077} &\numprint{1676} & \textbf{\numprint{1666}} &\numprint{2534} & \textbf{\numprint{2529}} \\ 
        wing\_nodal & \textbf{\numprint{1678}} & \numprint{1678} &\textbf{\numprint{3538}} & \numprint{3542} &\textbf{\numprint{5361}} & \numprint{5368} &\textbf{\numprint{8272}} & \numprint{8310} &\numprint{11939} & \textbf{\numprint{11828}} &\numprint{15967} & \textbf{\numprint{15874}} \\ 
        fe\_4elt2 & \textbf{\numprint{130}} & \numprint{130} &\textbf{\numprint{342}} & \numprint{342} &\textbf{\numprint{595}} & \numprint{596} &\textbf{\numprint{991}} & \numprint{994} &\textbf{\numprint{1599}} & \numprint{1613} &\textbf{\numprint{2485}} & \numprint{2503} \\ 
        vibrobox & \numprint{11538} & \textbf{\numprint{10310}} &\textbf{\numprint{18736}} & \numprint{18778} &\numprint{24204} & \textbf{\numprint{24170}} &\numprint{33065} & \textbf{\numprint{31514}} &\numprint{41312} & \textbf{\numprint{39512}} &\numprint{48184} & \textbf{\numprint{47651}} \\ 
        bcsstk29 & \textbf{\numprint{2818}} & \numprint{2818} &\textbf{\numprint{7971}} & \numprint{7983} &\textbf{\numprint{13717}} & \numprint{13816} &\numprint{22000} & \textbf{\numprint{21410}} &\numprint{34535} & \textbf{\numprint{34400}} &\numprint{55544} & \textbf{\numprint{55302}} \\ 
        4elt & \textbf{\numprint{137}} & \numprint{137} &\textbf{\numprint{319}} & \numprint{319} &\textbf{\numprint{522}} & \numprint{523} &\textbf{\numprint{906}} & \numprint{908} &\textbf{\numprint{1523}} & \numprint{1524} &\textbf{\numprint{2543}} & \numprint{2565} \\ 
        fe\_sphere & \textbf{\numprint{384}} & \numprint{384} &\textbf{\numprint{764}} & \numprint{764} &\textbf{\numprint{1152}} & \numprint{1152} &\textbf{\numprint{1698}} & \numprint{1704} &\numprint{2474} & \textbf{\numprint{2471}} &\numprint{3552} & \textbf{\numprint{3530}} \\ 
        cti & \textbf{\numprint{318}} & \numprint{318} &\textbf{\numprint{916}} & \numprint{916} &\textbf{\numprint{1714}} & \numprint{1714} &\textbf{\numprint{2746}} & \numprint{2758} &\textbf{\numprint{3994}} & \numprint{4011} &\textbf{\numprint{5579}} & \numprint{5675} \\ 
        memplus & \textbf{\numprint{5353}} & \numprint{5353} &\numprint{9375} & \textbf{\numprint{9362}} &\numprint{11662} & \textbf{\numprint{11624}} &\numprint{13088} & \textbf{\numprint{13001}} &\numprint{14617} & \textbf{\numprint{14107}} &\numprint{16997} & \textbf{\numprint{16259}} \\ 
        cs4 & \textbf{\numprint{360}} & \numprint{360} &\textbf{\numprint{917}} & \numprint{926} &\textbf{\numprint{1424}} & \numprint{1434} &\textbf{\numprint{2055}} & \numprint{2087} &\textbf{\numprint{2892}} & \numprint{2925} &\textbf{\numprint{4016}} & \numprint{4051} \\ 
        bcsstk30 & \textbf{\numprint{6251}} & \numprint{6251} &\textbf{\numprint{16399}} & \numprint{16497} &\textbf{\numprint{34137}} & \numprint{34275} &\textbf{\numprint{69592}} & \numprint{69763} &\numprint{113888} & \textbf{\numprint{113788}} &\numprint{173290} & \textbf{\numprint{171727}} \\ 
        bcsstk31 & \textbf{\numprint{2676}} & \numprint{2676} &\textbf{\numprint{7150}} & \numprint{7150} &\textbf{\numprint{12985}} & \numprint{13003} &\numprint{23299} & \textbf{\numprint{23232}} &\textbf{\numprint{37109}} & \numprint{37228} &\numprint{58143} & \textbf{\numprint{57953}} \\ 
        fe\_pwt & \textbf{\numprint{340}} & \numprint{340} &\textbf{\numprint{700}} & \numprint{700} &\textbf{\numprint{1410}} & \numprint{1411} &\textbf{\numprint{2773}} & \numprint{2776} &\textbf{\numprint{5460}} & \numprint{5488} &\textbf{\numprint{8124}} & \numprint{8205} \\ 
        bcsstk32 & \textbf{\numprint{4667}} & \numprint{4667} &\textbf{\numprint{8725}} & \numprint{8733} &\textbf{\numprint{19956}} & \numprint{19962} &\textbf{\numprint{35140}} & \numprint{35486} &\numprint{59716} & \textbf{\numprint{58966}} &\textbf{\numprint{91544}} & \numprint{91715} \\ 
        fe\_body & \textbf{\numprint{262}} & \numprint{262} &\textbf{\numprint{598}} & \numprint{598} &\numprint{1018} & \textbf{\numprint{1016}} &\textbf{\numprint{1708}} & \numprint{1734} &\textbf{\numprint{2738}} & \numprint{2810} &\textbf{\numprint{4643}} & \numprint{4799} \\ 
        t60k & \textbf{\numprint{71}} & \numprint{71} &\textbf{\numprint{203}} & \numprint{203} &\textbf{\numprint{449}} & \numprint{449} &\textbf{\numprint{793}} & \numprint{802} &\textbf{\numprint{1304}} & \numprint{1333} &\textbf{\numprint{2039}} & \numprint{2098} \\ 
        wing & \textbf{\numprint{773}} & \numprint{773} &\textbf{\numprint{1593}} & \numprint{1602} &\textbf{\numprint{2451}} & \numprint{2463} &\textbf{\numprint{3807}} & \numprint{3852} &\textbf{\numprint{5559}} & \numprint{5626} &\textbf{\numprint{7561}} & \numprint{7656} \\ 
        brack2 & \textbf{\numprint{684}} & \numprint{684} &\textbf{\numprint{2834}} & \numprint{2834} &\textbf{\numprint{6800}} & \numprint{6861} &\textbf{\numprint{11402}} & \numprint{11444} &\textbf{\numprint{17167}} & \numprint{17194} &\textbf{\numprint{25658}} & \numprint{25913} \\ 
        finan512 & \textbf{\numprint{162}} & \numprint{162} &\textbf{\numprint{324}} & \numprint{324} &\textbf{\numprint{648}} & \numprint{648} &\textbf{\numprint{1296}} & \numprint{1296} &\textbf{\numprint{2592}} & \numprint{2592} &\textbf{\numprint{10560}} & \numprint{10560} \\ 
        \hline
        fe\_tooth & \textbf{\numprint{3788}} & \numprint{3788} &\textbf{\numprint{6764}} & \numprint{6795} &\numprint{11287} & \textbf{\numprint{11274}} &\textbf{\numprint{17176}} & \numprint{17310} &\textbf{\numprint{24752}} & \numprint{24933} &\textbf{\numprint{34230}} & \numprint{34433} \\ 
        fe\_rotor & \textbf{\numprint{1959}} & \numprint{1959} &\textbf{\numprint{7118}} & \numprint{7126} &\textbf{\numprint{12445}} & \numprint{12472} &\textbf{\numprint{20076}} & \numprint{20112} &\textbf{\numprint{30664}} & \numprint{31233} &\textbf{\numprint{45053}} & \numprint{45911} \\ 
        598a & \textbf{\numprint{2367}} & \numprint{2367} &\textbf{\numprint{7816}} & \numprint{7838} &\textbf{\numprint{15613}} & \numprint{15722} &\textbf{\numprint{25563}} & \numprint{25686} &\textbf{\numprint{38346}} & \numprint{38627} &\textbf{\numprint{56153}} & \numprint{56179}\\ 
        fe\_ocean & \textbf{\numprint{311}} & \numprint{311} &\textbf{\numprint{1693}} & \numprint{1696} &\textbf{\numprint{3920}} & \numprint{3921} &\numprint{7657} & \textbf{\numprint{7631}} &\textbf{\numprint{12437}} & \numprint{12539} &\textbf{\numprint{19521}} & \numprint{19989} \\ 
        144 & \textbf{\numprint{6434}} & \numprint{6438} &\numprint{15203} & \textbf{\numprint{15078}} &\textbf{\numprint{25092}} & \numprint{25109} &\textbf{\numprint{37730}} & \numprint{37762} &\textbf{\numprint{55941}} & \numprint{56356} &\numprint{78636} & \textbf{\numprint{78559}}\\ 
        wave & \textbf{\numprint{8591}} & \numprint{8594} &\textbf{\numprint{16665}} & \numprint{16668} &\numprint{28506} & \textbf{\numprint{28495}} &\textbf{\numprint{42259}} & \numprint{42295} &\textbf{\numprint{60731}} & \numprint{61722} &\textbf{\numprint{84533}} & \numprint{85185} \\ 
        m14b & \textbf{\numprint{3823}} & \numprint{3823} &\textbf{\numprint{12948}} & \numprint{12948} &\textbf{\numprint{25390}} & \numprint{25520} &\textbf{\numprint{41778}} & \numprint{41997} &\numprint{65359} & \textbf{\numprint{65180}} &\textbf{\numprint{96519}} & \numprint{96802} \\ 
        auto & \textbf{\numprint{9673}} & \numprint{9683} &\textbf{\numprint{25789}} & \numprint{25836} &\textbf{\numprint{44785}} & \numprint{44832} &\textbf{\numprint{75719}} & \numprint{75778} &\textbf{\numprint{119157}} & \numprint{120086} &\textbf{\numprint{170989}} & \numprint{171535} \\ 
        \hline
        \end{tabular}
        \end{center} \caption{Computing partitions from scratch $\epsilon = 3$\%. In each $k$-column the results computed by KaFFPaE are on the left and the current Walshaw cuts are presented on the right side. }
        \end{table}
        \end{landscape}

\small
\begin{landscape}

\begin{table}[H]
\small
\begin{center}
\begin{tabular}{|l||r|r||r|r||r|r||r|r||r|r||r|r|}\hline
\small

Graph/$k$  & \multicolumn{2}{|c|}{2} & \multicolumn{2}{|c|}{4} & \multicolumn{2}{|c|}{8} & \multicolumn{2}{|c|}{16} & \multicolumn{2}{|c|}{32} & \multicolumn{2}{|c|}{64}\\

        \hline 
        add20 & \numprint{598} & \textbf{\numprint{546}} &\numprint{1169} & \textbf{\numprint{1149}} &\numprint{1689} & \textbf{\numprint{1675}} &\textbf{\numprint{2061}} & \numprint{2062} &\numprint{2411} & \textbf{\numprint{2387}} &\textbf{\numprint{2963}} & \numprint{3021} \\ 
        data & \numprint{182} & \textbf{\numprint{181}} &\textbf{\numprint{363}} & \numprint{363} &\textbf{\numprint{628}} & \numprint{628} &\numprint{1088} & \textbf{\numprint{1084}} &\numprint{1786} & \textbf{\numprint{1776}} &\numprint{2832} & \textbf{\numprint{2798}} \\ 
        3elt & \textbf{\numprint{87}} & \numprint{87} &\textbf{\numprint{197}} & \numprint{197} &\textbf{\numprint{329}} & \numprint{330} &\textbf{\numprint{557}} & \numprint{558} &\numprint{944} & \textbf{\numprint{942}} &\textbf{\numprint{1509}} & \numprint{1519} \\ 
        uk & \textbf{\numprint{18}} & \numprint{18} &\textbf{\numprint{39}} & \numprint{39} &\textbf{\numprint{75}} & \numprint{76} &\textbf{\numprint{137}} & \numprint{139} &\textbf{\numprint{237}} & \numprint{242} &\textbf{\numprint{395}} & \numprint{400} \\ 
        add32 & \textbf{\numprint{10}} & \numprint{10} &\textbf{\numprint{33}} & \numprint{33} &\textbf{\numprint{63}} & \numprint{63} &\textbf{\numprint{117}} & \numprint{117} &\textbf{\numprint{212}} & \numprint{212} &\textbf{\numprint{483}} & \numprint{486} \\ 
        bcsstk33 & \textbf{\numprint{9914}} & \numprint{9914} &\textbf{\numprint{20167}} & \numprint{20179} &\textbf{\numprint{33919}} & \numprint{33922} &\numprint{54333} & \textbf{\numprint{54296}} &\numprint{77457} & \textbf{\numprint{77101}} &\numprint{106903} & \textbf{\numprint{106827}} \\ 
        whitaker3 & \textbf{\numprint{126}} & \numprint{126} &\textbf{\numprint{377}} & \numprint{378} &\textbf{\numprint{644}} & \numprint{644} &\textbf{\numprint{1073}} & \numprint{1079} &\textbf{\numprint{1650}} & \numprint{1667} &\textbf{\numprint{2477}} & \numprint{2498} \\ 
        crack & \textbf{\numprint{182}} & \numprint{182} &\textbf{\numprint{360}} & \numprint{360} &\textbf{\numprint{666}} & \numprint{667} &\textbf{\numprint{1065}} & \numprint{1076} &\numprint{1661} & \textbf{\numprint{1655}} &\textbf{\numprint{2505}} & \numprint{2516} \\ 
        wing\_nodal & \numprint{1669} & \textbf{\numprint{1668}} &\textbf{\numprint{3521}} & \numprint{3522} &\textbf{\numprint{5341}} & \numprint{5345} &\textbf{\numprint{8241}} & \numprint{8264} &\textbf{\numprint{11793}} & \numprint{11828} &\numprint{15892} & \textbf{\numprint{15813}} \\ 
        fe\_4elt2 & \textbf{\numprint{130}} & \numprint{130} &\textbf{\numprint{335}} & \numprint{335} &\textbf{\numprint{578}} & \numprint{580} &\textbf{\numprint{983}} & \numprint{984} &\textbf{\numprint{1575}} & \numprint{1592} &\textbf{\numprint{2461}} & \numprint{2482} \\ 
        vibrobox & \numprint{11254} & \textbf{\numprint{10310}} &\textbf{\numprint{18690}} & \numprint{18696} &\textbf{\numprint{23924}} & \numprint{23930} &\numprint{32615} & \textbf{\numprint{31234}} &\numprint{40816} & \textbf{\numprint{39183}} &\numprint{47624} & \textbf{\numprint{47361}} \\ 
        bcsstk29 & \textbf{\numprint{2818}} & \numprint{2818} &\textbf{\numprint{7925}} & \numprint{7936} &\textbf{\numprint{13540}} & \numprint{13575} &\numprint{21459} & \textbf{\numprint{20924}} &\numprint{33851} & \textbf{\numprint{33817}} &\numprint{55029} & \textbf{\numprint{54895}} \\ 
        4elt & \textbf{\numprint{137}} & \numprint{137} &\textbf{\numprint{315}} & \numprint{315} &\textbf{\numprint{515}} & \numprint{515} &\textbf{\numprint{888}} & \numprint{895} &\textbf{\numprint{1504}} & \numprint{1516} &\textbf{\numprint{2514}} & \numprint{2546} \\ 
        fe\_sphere & \textbf{\numprint{384}} & \numprint{384} &\textbf{\numprint{762}} & \numprint{762} &\textbf{\numprint{1152}} & \numprint{1152} &\textbf{\numprint{1681}} & \numprint{1683} &\textbf{\numprint{2434}} & \numprint{2465} &\numprint{3528} & \textbf{\numprint{3522}} \\ 
        cti & \textbf{\numprint{318}} & \numprint{318} &\textbf{\numprint{889}} & \numprint{889} &\textbf{\numprint{1684}} & \numprint{1684} &\textbf{\numprint{2719}} & \numprint{2721} &\numprint{3927} & \textbf{\numprint{3920}} &\textbf{\numprint{5512}} & \numprint{5594} \\ 
        memplus & \numprint{5281} & \textbf{\numprint{5267}} &\textbf{\numprint{9292}} & \numprint{9297} &\numprint{11624} & \textbf{\numprint{11543}} &\numprint{13095} & \textbf{\numprint{13001}} &\numprint{14537} & \textbf{\numprint{14107}} &\numprint{16650} & \textbf{\numprint{16044}} \\ 
        cs4 & \textbf{\numprint{353}} & \numprint{353} &\textbf{\numprint{909}} & \numprint{912} &\textbf{\numprint{1420}} & \numprint{1431} &\textbf{\numprint{2043}} & \numprint{2079} &\textbf{\numprint{2866}} & \numprint{2919} &\textbf{\numprint{3973}} & \numprint{4012} \\ 
        bcsstk30 & \textbf{\numprint{6251}} & \numprint{6251} &\numprint{16189} & \textbf{\numprint{16186}} &\textbf{\numprint{34071}} & \numprint{34146} &\numprint{69337} & \textbf{\numprint{69288}} &\textbf{\numprint{112159}} & \numprint{113321} &\textbf{\numprint{170321}} & \numprint{170591} \\ 
        bcsstk31 & \textbf{\numprint{2669}} & \numprint{2670} &\textbf{\numprint{7086}} & \numprint{7088} &\textbf{\numprint{12853}} & \numprint{12865} &\textbf{\numprint{22871}} & \numprint{23104} &\textbf{\numprint{36502}} & \numprint{37228} &\numprint{57502} & \textbf{\numprint{56674}} \\ 
        fe\_pwt & \textbf{\numprint{340}} & \numprint{340} &\textbf{\numprint{700}} & \numprint{700} &\textbf{\numprint{1405}} & \numprint{1405} &\textbf{\numprint{2743}} & \numprint{2745} &\textbf{\numprint{5399}} & \numprint{5423} &\textbf{\numprint{7985}} & \numprint{8119} \\ 
        bcsstk32 & \textbf{\numprint{4622}} & \numprint{4622} &\textbf{\numprint{8441}} & \numprint{8441} &\textbf{\numprint{19411}} & \numprint{19601} &\textbf{\numprint{34481}} & \numprint{35014} &\textbf{\numprint{58395}} & \numprint{58966} &\numprint{90586} & \textbf{\numprint{89897}} \\ 
        fe\_body & \textbf{\numprint{262}} & \numprint{262} &\textbf{\numprint{588}} & \numprint{588} &\textbf{\numprint{1013}} & \numprint{1014} &\textbf{\numprint{1684}} & \numprint{1697} &\textbf{\numprint{2696}} & \numprint{2787} &\textbf{\numprint{4512}} & \numprint{4642} \\ 
        t60k & \textbf{\numprint{65}} & \numprint{65} &\textbf{\numprint{195}} & \numprint{195} &\textbf{\numprint{443}} & \numprint{445} &\textbf{\numprint{788}} & \numprint{796} &\textbf{\numprint{1299}} & \numprint{1329} &\textbf{\numprint{2021}} & \numprint{2089} \\ 
        wing & \textbf{\numprint{770}} & \numprint{770} &\textbf{\numprint{1590}} & \numprint{1593} &\textbf{\numprint{2440}} & \numprint{2452} &\textbf{\numprint{3775}} & \numprint{3832} &\textbf{\numprint{5538}} & \numprint{5564} &\textbf{\numprint{7567}} & \numprint{7611} \\ 
        brack2 & \textbf{\numprint{660}} & \numprint{660} &\textbf{\numprint{2731}} & \numprint{2731} &\textbf{\numprint{6592}} & \numprint{6611} &\textbf{\numprint{11193}} & \numprint{11232} &\textbf{\numprint{16919}} & \numprint{17112} &\textbf{\numprint{25598}} & \numprint{25805} \\ 
        finan512 & \textbf{\numprint{162}} & \numprint{162} &\textbf{\numprint{324}} & \numprint{324} &\textbf{\numprint{648}} & \numprint{648} &\textbf{\numprint{1296}} & \numprint{1296} &\textbf{\numprint{2592}} & \numprint{2592} &\textbf{\numprint{10560}} & \numprint{10560} \\ 
        \hline
        fe\_tooth & \textbf{\numprint{3773}} & \numprint{3773} &\textbf{\numprint{6688}} & \numprint{6714} &\textbf{\numprint{11154}} & \numprint{11185} &\textbf{\numprint{17070}} & \numprint{17215} &\textbf{\numprint{24733}} & \numprint{24933} &\textbf{\numprint{34320}} & \numprint{34433} \\ 
        fe\_rotor & \textbf{\numprint{1940}} & \numprint{1940} &\textbf{\numprint{6899}} & \numprint{6940} &\textbf{\numprint{12309}} & \numprint{12347} &\textbf{\numprint{19680}} & \numprint{19932} &\textbf{\numprint{30356}} & \numprint{30974} &\textbf{\numprint{45131}} & \numprint{45911} \\ 
        598a & \textbf{\numprint{2336}} & \numprint{2336} &\textbf{\numprint{7728}} & \numprint{7735} &\textbf{\numprint{15414}} & \numprint{15483} &\textbf{\numprint{25450}} & \numprint{25533} &\textbf{\numprint{38476}} & \numprint{38550} &\numprint{56377} & \textbf{\numprint{56179}} \\ 
        fe\_ocean & \textbf{\numprint{311}} & \numprint{311} &\textbf{\numprint{1686}} & \numprint{1686} &\textbf{\numprint{3893}} & \numprint{3902} &\textbf{\numprint{7385}} & \numprint{7412} &\textbf{\numprint{12211}} & \numprint{12362} &\textbf{\numprint{19400}} & \numprint{19727} \\ 
        144 & \textbf{\numprint{6357}} & \numprint{6359} &\numprint{15004} & \textbf{\numprint{14982}} &\numprint{25030} & \textbf{\numprint{24767}} &\numprint{37419} & \textbf{\numprint{37122}} &\textbf{\numprint{55460}} & \numprint{55984} &\textbf{\numprint{77430}} & \numprint{78069} \\ 
        wave & \textbf{\numprint{8524}} & \numprint{8533} &\numprint{16558} & \textbf{\numprint{16533}} &\textbf{\numprint{28489}} & \numprint{28492} &\textbf{\numprint{42084}} & \numprint{42134} &\textbf{\numprint{60537}} & \numprint{61280} &\textbf{\numprint{83413}} & \numprint{84236} \\ 
        m14b & \textbf{\numprint{3802}} & \numprint{3802} &\textbf{\numprint{12945}} & \numprint{12945} &\numprint{25154} & \textbf{\numprint{25143}} &\textbf{\numprint{41465}} & \numprint{41536} &\numprint{65237} & \textbf{\numprint{65077}} &\textbf{\numprint{96257}} & \numprint{96559} \\ 
        auto & \textbf{\numprint{9450}} & \numprint{9450} &\textbf{\numprint{25271}} & \numprint{25301} &\textbf{\numprint{44206}} & \numprint{44346} &\numprint{74636} & \textbf{\numprint{74561}} &\numprint{119294} & \textbf{\numprint{119111}} &\textbf{\numprint{169835}} & \numprint{171329} \\ 
        \hline
        \end{tabular}
        \end{center} \caption{Computing partitions from scratch $\epsilon = 5$\%. In each $k$-column the results computed by KaFFPaE are on the left and the current Walshaw cuts are presented on the right side. }
        \end{table}
        \end{landscape}

\end{appendix}
\end{document}